# Clause/Term Resolution and Learning in the Evaluation of Quantified Boolean Formulas

**Enrico Giunchiglia**                                        GIUNCHIGLIA@UNIGE.IT
**Massimo Narizzano**                                         MOX@DIST.UNIGE.IT
**Armando Tacchella**                                         TAC@DIST.UNIGE.IT
*DIST - Università di Genova*
*Viale Causa 13, 16145 Genova, Italy*

## Abstract

Resolution is the rule of inference at the basis of most procedures for automated reasoning. In these procedures, the input formula is first translated into an equisatisfiable formula in conjunctive normal form (CNF) and then represented as a set of clauses. Deduction starts by inferring new clauses by resolution, and goes on until the empty clause is generated or satisfiability of the set of clauses is proven, e.g., because no new clauses can be generated.

In this paper, we restrict our attention to the problem of evaluating Quantified Boolean Formulas (QBFs). In this setting, the above outlined deduction process is known to be sound and complete if given a formula in CNF and if a form of resolution, called "Q-resolution", is used. We introduce Q-resolution on terms, to be used for formulas in disjunctive normal form. We show that the computation performed by most of the available procedures for QBFs –based on the Davis-Logemann-Loveland procedure (*DLL*) for propositional satisfiability– corresponds to a tree in which Q-resolution on terms and clauses alternate. This poses the theoretical bases for the introduction of learning, corresponding to recording Q-resolution formulas associated with the nodes of the tree. We discuss the problems related to the introduction of learning in *DLL* based procedures, and present solutions extending state-of-the-art proposals coming from the literature on propositional satisfiability. Finally, we show that our *DLL* based solver extended with learning, performs significantly better on benchmarks used in the 2003 QBF solvers comparative evaluation.

## 1. Introduction

Resolution (Robinson, 1965) is the rule of inference at the basis of most procedures for automated reasoning (see, e.g., Fermüller, Leitsch, Hustadt, & Tammet, 2001; Bachmair & Ganzinger, 2001). In these procedures, the input formula is first translated into an equisatisfiable formula in conjunctive normal form (CNF) and then represented as a set of clauses. Deduction starts by inferring new clauses by resolution, and goes on until the empty clause is generated or satisfiability of the set of clauses is proven, e.g., because no new clauses can be generated. Here we restrict our attention to the problem of evaluating Quantified Boolean Formulas (QBFs). In this setting, the above outlined deduction process is known to be sound and complete if given a formula in CNF and if a form of resolution, called "Q-resolution", is used (Kleine-Büning, Karpinski, & Flögel, 1995). However, most of the available decision procedures for QBFs are based on and extend the Davis-Logemann-Loveland procedure (*DLL*) (Davis, Logemann, & Loveland, 1962) for propositional satisfiability (SAT). In the propositional case, it is well known that the computation performed by *DLL* corresponds to a specific form of resolution called "regular tree resolution" (see, e.g., Urquhart, 1995).

In this paper we introduce Q-resolution on terms, to be used for formulas in disjunctive normal form. We show that the computation performed by *DLL* based decision procedures for QBFs corresponds to a tree in which Q-resolution on terms and clauses alternate. Such correspondence poses the theoretical bases for the introduction of learning, corresponding to recording Q-resolution formulas associated with the nodes of the tree. In particular, recording Q-resolutions on clauses generalizes the popular "nogood" learning from constraint satisfaction and SAT literatures (see, e.g., Dechter, 1990; Bayardo, Jr. & Schrag, 1997): Each nogood corresponds to a set of assignments falsifying the input formula, and it is useful for pruning assignments to the existential variables. Recording Q-resolutions on terms corresponds to "good" learning: Each good corresponds to a set of assignments satisfying the input formula, and it is useful for pruning assignments to the universal variables. We discuss the problems related to the introduction of learning in *DLL* based decision procedures for QBFs, and present solutions extending state-of-the-art proposals coming from the literature on SAT. To show the effectiveness of learning for the QBFs evaluation problem, we have implemented it in QUBE, a state-of-the-art QBF solver. Using QUBE, we have done some experimental tests on several real-world QBFs, corresponding to planning (Rintanen, 1999; Castellini, Giunchiglia, & Tacchella, 2003) and circuit verification (Scholl & Becker, 2001; Abdelwaheb & Basin, 2000) problems, which are our two primary application domains of interest. The results witness the effectiveness of learning.

The paper is structured as follows. We first review the basics of Quantified Boolean Logic, at the same time introducing some terminology and notation that will be used throughout the paper. In Section 3, we introduce clause and term resolution, and their relation to *DLL* based decision procedures for QBFs. Then, in Section 4, we introduce both nogood and good learning, and then we show how they can be effectively integrated in *DLL* based decision procedures for QBFs. The implementation and the experimental results are presented in Section 5. The paper ends with the conclusions and some related work.

This paper builds on and extends in many ways our AAAI paper (Giunchiglia, Narizzano, & Tacchella, 2002). With respect to that paper, here (*i*) we introduce clause and term resolution; (*ii*) we show the correspondence between clause/term Q-resolution and the computation tree searched by DLL based decision procedures; (*iii*) on the basis of such correspondence, we extend the basic backtracking search procedure, first with backjumping and then with learning, and we prove their soundness and completeness; (*iv*) we discuss the implementation in QUBE providing many more details, and (*v*) we present the results of a much broader and detailed experimental analysis.

From here on, we simply write "resolution" for "Q-resolution".

## 2. Quantified Boolean Logic

Consider a set P of symbols. A *variable* is an element of P. A *literal* is a variable or the negation of a variable. In the following, for any literal $l$,

- $|l|$ is the variable occurring in $l$; and

- $\bar{l}$ is the negation of $l$ if $l$ is a variable, and it is $|l|$ otherwise.

For the sake of simplicity, we consider only formulas in negation normal form (NNF). Thus, for us, a *propositional formula* is a combination of literals using the $k$-ary ($k \geq 0$) connectives $\wedge$ (for conjunctions) and $\vee$ (for disjunctions). In the following, we use TRUE and FALSE as abbreviations for the empty conjunction and the empty disjunction respectively.

A *QBF* is an expression of the form

$$\varphi = Q_1 z_1 Q_2 z_2 \ldots Q_n z_n \Phi \qquad (n \geq 0) \tag{1}$$

where

- every $Q_i$ ($1 \leq i \leq n$) is a quantifier, either existential $\exists$ or universal $\forall$,

- $z_1, \ldots, z_n$ are distinct variables, and

- $\Phi$ is a propositional formula in $z_1, \ldots, z_n$.

For example,

$$\exists x_1 \forall y \exists x_2 ((\overline{x}_1 \vee \overline{y} \vee x_2) \wedge (\overline{y} \vee \overline{x}_2) \wedge (x_2 \vee ((x_1 \vee \overline{y}) \wedge (y \vee x_2)))) \tag{2}$$

is a QBF.

In (1), $Q_1 z_1 \ldots Q_n z_n$ is the *prefix* and $\Phi$ is the *matrix*. We also say that a literal $l$ is *existential* if $\exists |l|$ belongs to the prefix, and it is *universal* otherwise. Finally, in (1), we define

- the *level of a variable* $z_i$, to be 1 + the number of expressions $Q_j z_j Q_{j+1} z_{j+1}$ in the prefix with $j \geq i$ and $Q_j \neq Q_{j+1}$;

- the *level of a literal* $l$, to be the level of $|l|$.

For example, in (2) $x_2$ is existential and has level 1, $\overline{y}$ is universal and has level 2, $\overline{x}_1$ is existential and has level 3.

The *value* or *semantics* of a QBF $\varphi$ can be defined recursively as follows:

1. If the prefix is empty, then $\varphi$ is evaluated according to the truth tables of propositional logic.

2. If $\varphi$ is $\exists x \psi$, $\varphi$ is true if and only if $\varphi_x$ is true or $\varphi_{\overline{x}}$ is true.

3. If $\varphi$ is $\forall y \psi$, $\varphi$ is true if and only if both $\varphi_y$ and $\varphi_{\overline{y}}$ are true.

If $\varphi$ is (1) and $l$ is a literal with $|l| = z_i$, $\varphi_l$ is the QBF

- whose matrix is obtained from $\Phi$ by substituting

  - $z_i$ with TRUE and $\overline{z}_i$ with FALSE if $l = z_i$, and
  - $z_i$ with FALSE and $\overline{z}_i$ with TRUE if $l = \overline{z}_i$.

- whose prefix is $Q_1 z_1 Q_2 z_2 \ldots Q_{i-1} z_{i-1} Q_{i+1} z_{i+1} \ldots Q_n z_n$.

It is easy to see that if $\varphi$ is a QBF without universal quantifiers, the problem of determining the value of $\varphi$ reduces to the SAT problem.

Two QBFs are *equivalent* if they are either both true or both false.

## 3. Resolution and *DLL* Based Decision Procedures for QBFs

In this section we first introduce clause/term resolution and DLL based decision procedures for QBFs, and then we show the correspondence between the two.

### 3.1 Clause and Term Resolution

According to our definition of QBF, the matrix can be any combination of conjunctions and disjunctions of literals. However, using common clause form transformations based on renaming —first used by Tseitin (1970)—, it is possible to perform a linear time conversion from an arbitrary QBF into an equivalent one with the matrix in conjunctive normal form (CNF). These conversions are based on the fact that any QBF (1) is equivalent to

$$Q_1 z_1 Q_2 z_2 \ldots Q_n z_n \exists x ((\overline{x} \vee \Psi) \wedge \Phi[x/\Psi]) \qquad (n \geq 0)$$

where

- $\Psi$ is a propositional formula but not a literal;

- $x$ is a variable distinct from $z_1, z_2, \ldots, z_n$; and

- $\Phi[x/\Psi]$ is the propositional formula obtained from $\Phi$ substituting one or more occurrences of $\Psi$ with $x$.

Thus, if $\Psi$ is

$$((x_1 \vee \overline{y}) \wedge (y \vee x_2))$$

then it follows that (2) is equivalent to

$$\exists x_1 \forall y \exists x_2 \exists x_3 ((\overline{x}_1 \vee \overline{y} \vee x_2) \wedge (\overline{y} \vee \overline{x}_2) \wedge (x_2 \vee x_3) \wedge (x_1 \vee \overline{y} \vee \overline{x}_3) \wedge (y \vee x_2 \vee \overline{x}_3)) \quad (3)$$

Thanks to such conversions, we can restrict our attention to QBFs with the matrix in CNF, and represent the matrix of each formula as a set of clauses to be interpreted conjunctively, where a *clause* is a finite set of literals to be interpreted disjunctively. Further, we assume that each clause is non-tautological and minimal. A clause is *tautological* if it contains both a variable and its negation. A clause $C$ is *minimal* if the literals in $C$ with minimum level are existential. The *minimal form of a clause $C$* is the clause obtained from $C$ by deleting the universal literals which cause $C$ to be non-minimal. For instance, in (4), all the clauses are non-tautological and minimal. Our assumption that clauses are non-tautological and minimal is not a restriction, as the following theorem states.

**Theorem 1** *Let $\varphi$ be a QBF with the matrix in CNF. Let $\varphi'$ be the QBF obtained from $\varphi$ by*

1. *eliminating tautological clauses; and*

2. *replacing each non-tautological and non-minimal clause with its minimal form.*

*$\varphi$ and $\varphi'$ are equivalent.*

**Proof.** Clearly, tautological clauses can be eliminated from $\varphi$ and the result is an equivalent QBF. Let $C = \{l_1, \ldots, l_n, l_{n+1}, \ldots, l_m\}$ be a non-tautological and non-minimal clause in $\varphi$ in which $l_{n+1}, \ldots, l_m$ are the universal literals in $C \setminus min(C)$ ($0 \leq n < m$). Further, without loss of generality, we assume that the level of $l_i$ is less than or equal to the level of $l_{i+1}$, $1 \leq i < m$. Then, $\varphi$ has the form ($p \geq m$)

$$\ldots Q_1|l_1| \ldots \exists |l_n| \ldots \forall |l_{n+1}| \ldots \forall |l_m| Q_{m+1} z_{m+1} \ldots Q_p z_p \{\{l_1, \ldots, l_n, l_{n+1}, \ldots, l_m\}, \ldots\},$$

standing for

$$\ldots Q_1|l_1| \ldots \exists |l_n| \ldots \forall |l_{n+1}| \ldots \forall |l_m| Q_{m+1} z_{m+1} \ldots Q_p z_p ((l_1 \vee \ldots \vee l_n \vee l_{n+1} \vee \ldots \vee l_m) \wedge \Phi).$$

Then, by applying standard rules for quantifiers, $\varphi$ can be rewritten as

$$\ldots Q_1|l_1| \ldots \exists |l_n| \ldots \forall |l_{n+1}| \ldots \forall |l_m| ((l_1 \vee \ldots \vee l_n \vee l_{n+1} \vee \ldots \vee l_m) \wedge Q_{m+1} z_{m+1} \ldots Q_p z_p \Phi),$$

equivalent to

$$\ldots Q_1|l_1| \ldots \exists |l_n| \ldots \forall |l_{n+1}| \ldots (\forall |l_m| (l_1 \vee \ldots \vee l_n \vee l_{n+1} \vee \ldots \vee l_m) \wedge \forall |l_m| Q_{m+1} z_{m+1} \ldots Q_p z_p \Phi),$$

equivalent to

$$\ldots Q_1|l_1| \ldots \exists |l_n| \ldots \forall |l_{n+1}| \ldots ((l_1 \vee \ldots \vee l_n \vee l_{n+1} \vee \ldots \vee l_{m-1}) \wedge \forall |l_m| Q_{m+1} z_{m+1} \ldots Q_p z_p \Phi),$$

equivalent to

$$\ldots Q_1|l_1| \ldots \exists |l_n| \ldots \forall |l_{n+1}| \ldots \forall |l_m| Q_{m+1} z_{m+1} \ldots Q_p z_p ((l_1 \vee \ldots \vee l_n \vee l_{n+1} \vee \ldots \vee l_{m-1}) \wedge \Phi),$$

i.e., the QBF obtained from $\varphi$ by deleting $l_m$ from the clause $C$. By iterating the above reasoning process, all the literals in $C \setminus min(C)$ can be eliminated from $C$, and hence the thesis. $\square$

From here on, a *QBF is in CNF* if and only if the matrix is a conjunction of clauses, and each clause is both minimal and non-tautological. If we represent the matrix of a QBF as a set of clauses,

- the empty clause $\{\}$ stands for False;

- the empty set of clauses $\{\}$ stands for True;

- the formula $\{\{\}\}$ is equivalent to False;

- the QBF (3) is written as

$$\exists x_1 \forall y \exists x_2 \exists x_3 \{\{\overline{x}_1, \overline{y}, x_2\}, \{\overline{y}, \overline{x}_2\}, \{x_2, x_3\}, \{x_1, \overline{y}, \overline{x}_3\}, \{y, x_2, \overline{x}_3\}\}. \quad (4)$$

Clause resolution (Kleine-Büning et al., 1995) is similar to an ordinary resolution where only existential literals can be matched. More precisely, *clause resolution (on a literal $l$)* is the rule

$$\frac{C_1 \qquad C_2}{min(C)} \quad (5)$$

where

| (c1) | $\{\overline{x}_1, \overline{y}, x_2\}$ | Input formula | (c5) | $\{\overline{x}_1\}$ | From (c1), (c2) |
|---|---|---|---|---|---|
| (c2) | $\{\overline{y}, \overline{x}_2\}$ | Input formula | (c6) | $\{x_3, \overline{y}\}$ | From (c2), (c3) |
| (c3) | $\{x_2, x_3\}$ | Input formula | (c7) | $\{x_1\}$ | From (c4), (c6) |
| (c4) | $\{x_1, \overline{y}, \overline{x}_3\}$ | Input formula | (c8) | $\{\}$ | From (c5), (c7) |

Table 1: A clause resolution deduction showing that (4) is false. The prefix is $\exists x_1 \forall y \exists x_2 \exists x_3$.

- $l$ is an existential literal;

- $C_1$, $C_2$ are two clauses such that $\{l, \overline{l}\} \subseteq (C_1 \cup C_2)$, and for no literal $l' \neq l$, $\{l', \overline{l'}\} \subseteq (C_1 \cup C_2)$;

- $C$ is $(C_1 \cup C_2) \setminus \{l, \overline{l}\}$.

$C_1$ and $C_2$ are the *antecedents*, and $min(C)$ is the *resolvent* of the rule.

**Theorem 2 ((Kleine-Büning et al., 1995))** *Clause resolution is a sound and complete proof system for deciding QBFs in CNF: a QBF in CNF is true if and only if the empty clause is not derivable by clause resolution.*

For instance, the fact that (4) is false follows from the deduction in Table 1.

Alternatively to the CNF conversion, we could have converted (2) into a QBF with the matrix in disjunctive normal form (DNF), again in linear time, on the basis that any QBF (1), is equivalent to

$$Q_1 z_1 Q_2 z_2 \ldots Q_n z_n \forall y ((\overline{y} \wedge \Psi) \vee \Phi[y/\Psi]) \qquad (n \geq 0),$$

assuming $\Psi$ is a propositional formula but not a literal, and that $y$ is a variable distinct from $z_1, z_2, \ldots, z_n$.

A simple recursive application of the above equivalence to (2) leads to the following equivalent QBF:

$$\begin{aligned}
\exists x_1 \forall y \exists x_2 \forall y_1 \forall y_2 \forall y_3 \forall y_4 \forall y_5 \forall y_6 ((y_1 \wedge y_2 \wedge y_3) \vee \\
(\overline{y}_1 \wedge \overline{x}_1) \vee (\overline{y}_1 \wedge \overline{y}) \vee (\overline{y}_1 \wedge x_2) \vee \\
(\overline{y}_2 \wedge \overline{y}) \vee (\overline{y}_2 \wedge \overline{x}_2) \vee \\
(\overline{y}_3 \wedge x_2) \vee (\overline{y}_3 \wedge y_4) \vee \\
(\overline{y}_4 \wedge y_5 \wedge y_6) \vee \\
(\overline{y}_5 \wedge x_1) \vee (\overline{y}_5 \wedge \overline{y}) \vee \\
(\overline{y}_6 \wedge y) \vee (\overline{y}_6 \wedge x_2)).
\end{aligned} \qquad (6)$$

Given a QBF with the matrix in DNF, we can represent the matrix as a set of terms to be interpreted disjunctively, where a *term* is a finite set of literals to be interpreted conjunctively. Further, we can assume that each term is non-contradictory and minimal. A term is *contradictory* if it contains both a variable and its negation. A term $T$ is *minimal* if the literals in $T$ with minimum level are universal. The *minimal form of a term $T$* is the term obtained from $T$ by deleting the existential literals which cause $T$ to be non-minimal. All the terms in (6) are non-contradictory and minimal. Analogously to what we have said before for QBFs in CNF, if $\varphi$ is a QBF in DNF then we can assume that all the terms are non-contradictory and minimal without loss of generality.

**Theorem 3** *Let $\varphi$ be a QBF with the matrix in DNF. Let $\varphi'$ be the QBF obtained from $\varphi$ by*

1. *eliminating contradictory terms; and*

2. *replacing each non-contradictory and non-minimal term with its minimal form.*

$\varphi$ *and* $\varphi'$ *are equivalent.*

**Proof.** Analogous to the proof of Theorem 1. □

As before, from here on, a *QBF is in DNF* if and only if the matrix is a disjunction of terms, and each term is both minimal and non-contradictory.

We can introduce *term resolution (on a literal $l$)* which consists of the rule

$$\frac{T_1 \qquad T_2}{min(T)}$$

where

- $l$ is an universal literal;

- $T_1$, $T_2$ are two terms such that $\{l, \bar{l}\} \subseteq (T_1 \cup T_2)$, and for no literal $l' \neq l$, $\{l', \bar{l'}\} \subseteq (T_1 \cup T_2)$;

- $T$ is $(T_1 \cup T_2) \setminus \{l, \bar{l}\}$.

$T_1$ and $T_2$ are the *antecedents*, and $min(T)$ is the *resolvent* of the rule.

**Theorem 4** *Term resolution is a sound and complete proof system for deciding QBFs in DNF: a QBF in DNF is true if and only if the empty term is derivable by term resolution.*

**Proof.** The fact that term resolution is a sound and complete proof system follows from the soundness and completeness of clause resolution.

Let $\Phi$ be a set of sets of literals, and $\varphi = Q_1 z_1 Q_2 z_2 \ldots Q_n z_n \Phi$ be a QBF in which $\Phi$ is interpreted as a set of clauses. Without loss of generality we can assume that each clause in $\Phi$ is non-tautological and minimal. Then the following chain of equivalences holds:

There exists a deduction $\Delta$ of the empty clause from $\varphi$ using clause resolution

if and only if

$\varphi$ is false

if and only if

The QBF $\overline{\varphi} = \overline{Q}_1 z_1 \overline{Q}_2 z_2 \ldots \overline{Q}_n z_n \Phi$ in which $\Phi$ is interpreted as a set of terms is true

if and only if

$\Delta$ is a deduction of the empty term from $\overline{\varphi}$ using term resolution.

In the above chain of equivalences, $\overline{Q}$ is $\exists$ if $Q = \forall$, and is $\forall$ if $Q = \exists$. □

As an example of a term resolution deduction of the empty term, consider the QBF $\overline{\varphi}$:

$$\forall x_1 \exists y \forall x_2 \forall x_3 ((\overline{x}_1 \wedge \overline{y} \wedge x_2) \vee (\overline{y} \wedge \overline{x}_2) \vee (x_2 \wedge x_3) \vee (x_1 \wedge \overline{y} \wedge \overline{x}_3) \vee (y \wedge x_2 \wedge \overline{x}_3))$$

i.e., the QBF obtained from (3) by simultaneously replacing $\forall$ with $\exists$, $\exists$ with $\forall$, $\wedge$ with $\vee$, and $\vee$ with $\wedge$. Then, the deduction in Table 1 is also a deduction of the empty term from $\overline{\varphi}$ using term resolution.

If the QBF $\varphi$ is not in DNF but in CNF then term resolution cannot be applied, and thus term resolution is not sufficient for proving the truth or falsity of $\varphi$. However, if we also have the following *model generation* rule

$$\frac{\Phi}{min(T)}$$

where

- $\Phi$ is the matrix of $\varphi$; and

- $T$ is a non-contradictory term such that for each clause $C \in \Phi$, $C \cap T \neq \emptyset$,

we get a sound and complete proof system for QBFs in CNF. Intuitively, the model generation rule allows us to start from the minimal form of terms which propositionally entail the matrix of the input formula.

**Theorem 5** *Term resolution and model generation is a sound and complete proof system for deciding QBFs in CNF: a QBF in CNF is true if and only if the empty term is derivable by term resolution and model generation.*

**Proof.** Given a QBF $\varphi$ in CNF with matrix $\Phi$, by the model generation rule we can derive a set $\Psi$ of terms of the form $min(T)$ such that

- each term $T$ is a non-contradictory and such that for each clause $C \in \Phi$, $C \cap T \neq \emptyset$; and

- the disjunction of all the terms in $\Psi$ is propositionally logically equivalent to $\Phi$.

Let $\varphi'$ be the QBF in DNF obtained by substituting $\Phi$ with $\Psi$ in $\varphi$. $\varphi$ and $\varphi'$ have the same value. Hence the thesis thanks to Theorem 4. $\square$

## 3.2 *DLL* Based Decision Procedures for QBFs

Given what we have said so far, an arbitrary QBF $\varphi$ can be converted (in linear time) into an equivalent QBF in CNF. Because of this, from here to the end of the paper, we restrict our attention to QBFs in such format. With this assumption, if $\varphi$ is (1) and $l$ is a literal with $|l| = z_i$, we redefine $\varphi_l$ to be the QBF

- whose matrix is obtained from $\Phi$ by removing the clauses $C$ with $l \in C$, and by removing $\bar{l}$ from the other clauses; and

- whose prefix is $Q_1 z_1 Q_2 z_2 \ldots Q_{i-1} z_{i-1} Q_{i+1} z_{i+1} \ldots Q_n z_n$.

Further, we extend the notation to sequence of literals: If $\mu = l_1; l_2; \ldots; l_m$ $(m \geq 0)$, $\varphi_\mu$ is defined as $(\ldots((\varphi_{l_1})_{l_2})\ldots)_{l_m}$.

Consider a QBF $\varphi$.

A simple procedure for determining the value of $\varphi$ starts with the empty assignment $\epsilon$ and recursively extends the current assignment $\mu$ with $z$ and/or $\overline{z}$, where $z$ is a heuristically chosen variable at the highest level in $\varphi_\mu$, until either the empty clause or the empty set of clauses are produced in $\varphi_\mu$. On the basis of the values of $\varphi_{\mu;z}$ and $\varphi_{\mu;\overline{z}}$, the value of $\varphi_\mu$ can be determined according to the semantics of QBFs. The value of $\varphi$ is the value of $\varphi_\epsilon$.

Cadoli, Giovanardi, Giovanardi and Schaerf (2002) introduced various improvements to this basic procedure.

The first improvement is that we can directly conclude about the value of $\varphi_\mu$ if the matrix of $\varphi_\mu$ contains a contradictory clause (Lemma 2.1 in Cadoli et al., 2002). A clause $C$ is *contradictory* if it contains no existential literal. An example of a contradictory clause is the empty clause.

The second improvement allows us to directly extend $\mu$ with $l$ if $l$ is unit or monotone in $\varphi_\mu$ (Lemmas 2.4, 2.5, 2.6 in Cadoli et al., 2002). In (1), a literal $l$ is:

- *Unit* if $l$ is existential and for some $m \geq 0$,

    - a clause $\{l, l_1, \ldots, l_m\}$ belongs to $\Phi$; and

    - each literal $l_i$ $(1 \leq i \leq m)$ is universal and has a level lower than the level of $l$.

- *Monotone* or *pure* if

    - either $l$ is existential, $\overline{l}$ does not belong to any clause in $\Phi$, and $l$ occurs in $\Phi$;

    - or $l$ is universal, $l$ does not belong to any clause in $\Phi$, and $\overline{l}$ occurs in $\Phi$.

For example, in a QBF of the form

$$\ldots \exists x_1 \forall y \exists x_2 \ldots \{\{x_1, y\}, \{x_2\}, \ldots\},$$

both $x_1$ and $x_2$ are unit. In the QBF

$$\forall y_1 \exists x_1 \forall y_2 \exists x_2 \{\{\overline{y}_1, y_2, x_2\}, \{x_1, \overline{y}_2, \overline{x}_2\}\},$$

the only monotone literals are $y_1$ and $x_1$.

With such improvements, the resulting procedure, called *Q-DLL*, is essentially the one presented in the work of Cadoli, Giovanardi, and Schaerf (1998), which extends *DLL* in order to deal with QBFs. Figure 1 is a simple, recursive presentation of it. In the figure, given a QBF $\varphi$,

1. FALSE is returned if a contradictory clause is in the matrix of $\varphi_\mu$ (line 1); otherwise

2. TRUE is returned if the matrix of $\varphi_\mu$ is empty (line 2); otherwise

3. at line 3, $\mu$ is recursively extended to $\mu; l$ if $l$ is unit (and we say that $l$ has been *assigned as unit*); otherwise

```
0  function Q-DLL(φ, μ)
1      if (⟨a contradictory clause is in the matrix of φ_μ⟩) return FALSE;
2      if (⟨the matrix of φ_μ is empty⟩) return TRUE;
3      if (⟨l is unit in φ_μ⟩) return Q-DLL(φ, μ; l);
4      if (⟨l is monotone in φ_μ⟩) return Q-DLL(φ, μ; l);
5      l := ⟨a literal at the highest level in φ_μ⟩;
6      if (⟨l is existential⟩) return Q-DLL(φ, μ; l) or Q-DLL(φ, μ; l̄);
7      else return Q-DLL(φ, μ; l) and Q-DLL(φ, μ; l̄).
```

Figure 1: The algorithm of *Q-DLL*.

4. at line 4, $\mu$ is recursively extended to $\mu; l$ if $l$ is monotone (and we say that $l$ has been *assigned as monotone*); otherwise

5. a literal $l$ at the highest level is chosen and

- If $l$ is existential (line 6), $\mu$ is extended to $\mu; l$ first (and we say that $l$ has been *assigned as left split*). If the result is FALSE, $\mu; \bar{l}$ is tried and returned (and in this case we say that $\bar{l}$ has been *assigned as right split*).

- Otherwise (line 7), $l$ is universal, $\mu$ is extended to $\mu; l$ first (and we say that $l$ has been *assigned as left split*). If the result is TRUE, $\mu; \bar{l}$ is tried and returned (and in this case we say that $\bar{l}$ has been *assigned as right split*).

**Theorem 6** *Q-DLL($\varphi, \epsilon$) returns* TRUE *if $\varphi$ is true, and* FALSE *otherwise.*

**Proof.** Trivial consequence of Lemmas 2.1, 2.4, 2.5, 2.6 in the work of Cadoli, Giovanardi, Giovanardi, and Schaerf (2002) and of the semantics of QBFs. □

Given what we have said so far, it is clear that *Q-DLL* evaluates $\varphi$ by generating a semantic tree (Robinson, 1968) in which each node corresponds to an invocation of *Q-DLL* and thus to an assignment $\mu$. For us,

- an *assignment (for a QBF $\varphi$)* is a possibly empty sequence $\mu = l_1; l_2; \ldots; l_m$ $(m \geq 0)$ of literals such that for each $l_i$ in $\mu$, $l_i$ is unit, or monotone, or at the highest level in $\varphi_{l_1; l_2; \ldots; l_{i-1}}$;

- the *(semantic) tree* representing a run of *Q-DLL* on $\varphi$ is the tree

    - having a node $\mu$ for each call to *Q-DLL($\varphi, \mu$)*; and

    - an edge connecting any two nodes $\mu$ and $\mu; l$, where $l$ is a literal.

Any tree representing a run of *Q-DLL* has at least the node $\epsilon$.

As an example of a run of *Q-DLL*, consider the QBF (4). For simplicity, assume that the literal returned at line 5 in Figure 1 is the negation of the first variable in the prefix which occurs in the matrix of the QBF under consideration. Then, the tree searched by *Q-DLL* when $\varphi$ is (4) can be represented as in Figure 2. In the figure:

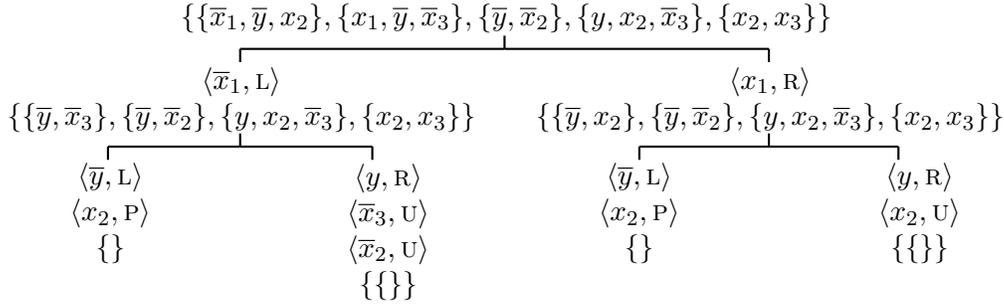

Figure 2: The tree generated by *Q-DLL* for (4). The matrix of (4) is shown at the root node, and the prefix is $\exists x_1 \forall y \exists x_2 \exists x_3$. U, P, L, R stand for "unit", "pure", "left split", "right split" respectively, and have the obvious meaning.

- Each node is labeled with the literal assigned by *Q-DLL* in order to extend the assignment built so far. Thus, the assignment corresponding to a node is the sequence of labels in the path from the root to the node. For instance, the assignment corresponding to the node with label $\overline{x}_3$ is $\overline{x}_1; y; \overline{x}_3$.

- When literals are assigned as unit or monotone, the corresponding nodes are aligned one below the other. Further for each assigned literal $l$, we also show whether $l$ has been assigned as unit, monotone, left or right split by marking it as U, P, L, R respectively.

- When $l$ has been assigned as a left or right split, we also show the matrix of $\varphi_{\mu;l}$, where $\mu$ is the sequence of literals assigned before $l$.

- When the node $\mu$ is a leaf, then the matrix of $\varphi_\mu$ is either empty (in which case we write "{}" below the node), or it contains a contradictory clause (in which case we write "{{}}" below the node).

Considering Figure 2, it is easy to see that *Q-DLL* would correctly return FALSE, meaning that (4) (and thus also (2)) is false.

### 3.3 Resolution and *DLL* Based Decision Procedures for QBFs

The well known correspondence in SAT between semantic trees and resolution (see, e.g., Urquhart, 1995) gives us the starting point for our analysis, aimed to establish a correspondence between *Q-DLL* and clause/term resolution.

Consider a QBF $\varphi$. Let $\Pi$ be the tree explored by *Q-DLL* for evaluating $\varphi$.

For the time being, assume that we are dealing with a SAT problem, i.e., $\varphi$ does not contain universal quantifiers. Then, *Q-DLL* reduces to *DLL*, and if $\varphi$ is false then we can use $\Pi$ to generate a clause resolution deduction of the empty clause from $\varphi$. The basic idea is to associate with each node $\mu$ of $\Pi$ a clause $C$ which is $\mu$-falsified, i.e., such that for each literal $l \in C$, $\bar{l}$ is in $\mu$. (We say that a literal $l$ *is in* or *has been assigned by* $l_1; \ldots; l_m$ if and only if $l \in \{l_1, \ldots, l_m\}$). More precisely:

- With every leaf $\mu$ of $\Pi$, we associate an arbitrarily selected clause in the matrix of $\varphi$ which is $\mu$-falsified. At least one such clause exists because $\varphi_\mu$ contains the empty clause.

- If $C$ is the clause associated with a node $\mu; l$, then

  1. If $\bar{l} \notin C$ then $C$ is also the clause associated with $\mu$. Notice that if $l$ is monotone in $\varphi_\mu$ then $\bar{l} \notin C$.

  2. If $\bar{l} \in C$ and $l$ is unit in $\varphi_\mu$ then the clause associated with $\mu$ is the resolvent of $C$ and an arbitrarily selected clause of $\varphi$ which causes $l$ to be unit in $\varphi_\mu$.

  3. If $\bar{l} \in C$ and $l$ is not unit in $\varphi_\mu$ then we have to consider the clause $C'$ associated with the node $\mu; \bar{l}$. If $l \notin C'$ then $C'$ is the clause associated with $\mu$ (as in the first case). If $l \in C'$, the clause associated with $\mu$ is the resolvent of $C$ and $C'$.

**Lemma 1** *Let $\varphi$ be a QBF without universal quantifiers. Let $\Pi$ be the tree searched by Q-DLL$(\varphi, \epsilon)$. Let $\mu$ be an assignment for $\varphi$. If $\varphi_\mu$ is false, then the clause associated with the node $\mu$ of $\Pi$*

- *is $\mu$-falsified; and*

- *does not contain existential literals whose negation has been assigned as monotone in $\mu$.*

**Proof.** Let $S$ be the set of assignments in $\Pi$ which extend $\mu$. Clearly, for each assignment $\mu' \in S$, $\varphi_{\mu'}$ is false ($\varphi$ does not contain universal quantifiers). On $S$, we define the partial order relation $\succeq$ according to which two assignments $\mu'$ and $\mu''$ in $S$ are such that $\mu' \succeq \mu''$ if and only if $\mu'$ extends $\mu''$. Clearly $\succeq$ is well founded and the minimal elements are the assignments extending $\mu$ and corresponding to the leaves of $\Pi$.

If $\mu'$ extends $\mu$ and is a leaf of $\Pi$, then $\varphi_{\mu'}$ contains a contradictory clause $C$. Since $\varphi$ does not contain universal quantifiers, $C$ is $\mu'$-falsified and is associated with the node $\mu'$. Clearly, $C$ does not contain existential literals whose negation has been assigned as monotone.

By induction hypothesis, for each assignment $\mu' = \mu''; l \succeq \mu''$ in $S$ we have a $\mu'$-falsified clause not containing existential literals whose negation has been assigned as monotone. We have to show the thesis for $\mu''$. There are three cases:

1. $l$ has been assigned as unit. Let $C_1$ be the clause associated with $\mu''; l$. By induction hypothesis, the thesis holds for $C_1$. If $C_1$ does not contain $\bar{l}$, the thesis trivially follows. Otherwise, the clause associated with $\mu''$ is the resolvent $C$ of $C_1$ with a clause $C_2$ that causes $l$ to be unit in $\varphi_{\mu''}$. $C_2$ is $\mu''; \bar{l}$-falsified and it does not contain existential literals whose negation has been assigned as monotone. $C = C_1 \cup C_2 \setminus \{l, \bar{l}\}$ and thus the thesis trivially holds.

2. $l$ has been assigned as monotone. In this case the clause $C$ associated with $\mu''$ is the same clause associated with $\mu''; l$. By induction hypothesis $C$ does not contain $\bar{l}$ and thus $C$ is $\mu''$-falsified.

3. $l$ is a split. In this case we have a clause $C_1$ associated with $\mu''; l$ and a clause $C_2$ associated with $\mu''; \bar{l}$. The thesis holds for both $C_1$ and $C_2$ by induction hypothesis. If $C_1$ does not contain $\bar{l}$, then the clause associated with $\mu''$ is $C_1$ and the thesis trivially holds. Otherwise, if $C_2$ does not contain $l$, then the clause associated with $\mu''$ is $C_2$ and again the thesis trivially holds. Otherwise, the clause associated with $\mu''$ is $C_1 \cup C_2 \setminus \{l, \bar{l}\}$ and again the thesis trivially holds.

$\square$

**Theorem 7** *Let $\varphi$ be a false QBF without universal quantifiers. The tree searched by Q-DLL$(\varphi, \epsilon)$ corresponds to a clause resolution deduction of the empty clause.*

**Proof.** Let $\Delta$ be a sequence of clauses obtained by listing the clauses in the matrix of $\varphi$ according to an arbitrary order, followed by the clauses associated with the internal nodes of the tree $\Pi$ searched by Q-DLL$(\varphi, \epsilon)$, assuming $\Pi$ is visited in post order. Clearly, $\Delta$ is a deduction. $\Delta$ is a deduction of the empty clause because the node $\epsilon$ has an associated $\epsilon$-falsified clause (Lemma 1), i.e., the empty clause. $\square$

The theorem points out the close correspondence between the computation of *Q-DLL* and clause resolution, assuming the input formula is false and that it does not contain universal quantifiers. If the input formula does not contain universal quantifiers but is true, still the tree explored by *Q-DLL* before generating the path ending with the empty matrix corresponds to a sequence of clause resolutions, one for each maximal subtree whose leaves $\mu$ are such that $\varphi_\mu$ contains an empty clause.

If we no longer assume that the input formula $\varphi$ does not contain universal quantifiers, and consider the case in which $\varphi$ is an arbitrary QBF, the situation gets more complicated, also because of the possibility of assigning unit literals which are not at the highest level. So, we now assume that if a literal $l$ is assigned as unit at a node $\mu$, then $l$ is at the highest level in $\varphi_\mu$.

Then, if the input formula $\varphi$ is false, we can again use the tree $\Pi$ searched by *Q-DLL* to generate a clause resolution deduction of the empty clause. The construction is analogous to the one described before. The only difference is that we have to restrict our attention to the *minimal false subtree* of $\Pi$, i.e., the tree obtained from $\Pi$ by deleting the subtrees starting with a left split on a universal literal: These subtrees are originated from "wrong choices" when deciding which branch to explore first. In the minimal false subtree $\Pi'$ of $\Pi$, all the leaves terminate with the empty clause, and we can associate with each node of $\Pi'$ a clause exactly in the same way described above for the SAT case. For instance, if $\varphi$ is (4), then *Q-DLL* assigns unit literals only when they are at the highest level. Figure 3 shows the minimal false subtree of *Q-DLL*'s computation, and the associated clause resolution deduction of the empty clause. In the figure,

- the clause associated with each node is written in red and to the right of the node itself;

- when a node corresponds to the assignment of a unit literal $l$, a clause of $\varphi$ which causes $l$ to be unit at that node (used in the corresponding clause resolution) is written in red and to the left of the node.

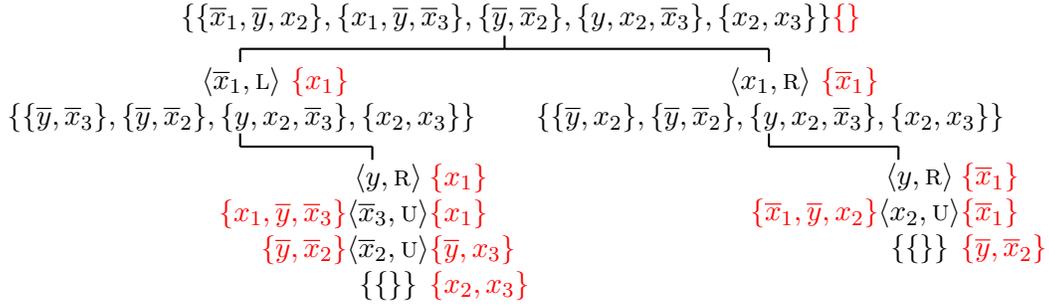

Figure 3: The clause resolution corresponding to the tree generated by *Q-DLL* for (4). The prefix is $\exists x_1 \forall y \exists x_2 \exists x_3$.

**Lemma 2** *Let $\varphi$ be a false QBF. Let $\Pi$ be the minimal false subtree of the tree searched by Q-DLL$(\varphi, \epsilon)$ and assume that for each node $\mu; l$ in $\Pi$, if $l$ is unit in $\varphi_\mu$ then $l$ is also at the highest level in $\varphi_\mu$. Let $\mu$ be an assignment for $\varphi$. If $\varphi_\mu$ is false, then the clause associated with the node $\mu$ of $\Pi$*

- *is $\mu$-falsified; and*

- *does not contain existential literals whose negation has been assigned as monotone in $\mu$.*

**Proof.** Trivial extension of the proof of Lemma 1. The assumption that for each node $\mu; l$ in $\Pi$, if $l$ is unit in $\varphi_\mu$ then $l$ is at the highest level in $\varphi_\mu$, ensures that each clause associated with a node $\mu$ of $\Pi$ is $\mu$-falsified. $\qquad\square$

**Theorem 8** *Let $\varphi$ be a false QBF. Let $\Pi$ be the minimal false subtree of the tree searched by Q-DLL$(\varphi, \epsilon)$ and assume that for each node $\mu; l$ in $\Pi$, if $l$ is unit in $\varphi_\mu$ then $l$ is also at the highest level in $\varphi_\mu$. Then $\Pi$ corresponds to a clause resolution deduction of the empty clause.*

**Proof.** Given Lemma 2, the proof is analogous to the one of Theorem 7. $\qquad\square$

Regardless of whether the input formula is true or false, the tree explored by *Q-DLL* may contain (exponentially many) subtrees whose nodes $\mu$ are such that $\varphi_\mu$ is false. The procedure described above, allows us to associate a clause resolution deduction with each of such subtrees.

If the input formula $\varphi$ is true, the situation is simpler because so far we do not have "unit universal literals", and we can use the tree $\Pi$ searched by *Q-DLL* to generate a deduction of the empty term from $\varphi$. Intuitively, the process is analogous to the one described when $\varphi$ is false, except that the leaves of our term resolution deduction are terms corresponding to the assignments computed by *Q-DLL* and entailing the matrix of $\varphi$. In details:

- First, we have to restrict our attention to the *minimal true subtree* of $\Pi$, i.e., the tree obtained from $\Pi$ by deleting the subtrees starting with a left split on an existential literal: Analogously to the the case in which $\varphi$ is false, each leaf in a minimal true subtree of $\Pi$ terminates with the empty matrix.

- Second, we associate with each node $\mu$ a term, represented as a set, as follows:

  - The term associated with each leaf is a minimal term $min(T)$ in which $T^{1}$

    1. does not contain universal literals assigned as monotone,
    2. has to propositionally entail the matrix, i.e., for each clause $C$ in the matrix of $\varphi$, $T \cap C \neq \emptyset$, and
    3. has to be a subset of the literals in $\mu$, i.e., $T \subseteq \{l : l \text{ is in } \mu\}$.

  - If $T$ is the term associated with a node $\mu;l$, then

    1. If $l \notin T$ then $T$ is the term associated with the node $\mu$. Notice that if $l$ is either existential or both universal and monotone in $\varphi_\mu$, then $l \notin T$.
    2. If $l \in T$ then we have to consider also the term $T'$ associated with the node $\mu;\bar{l}$. If $\bar{l} \notin T'$ then $T'$ is the term associated with $\mu$ (as in the first case). If $\bar{l} \in T'$, the term associated with $\mu$ is the resolvent of $T$ and $T'$.

It is easy to see that the term $T$ associated with a node $\mu$ is $\mu$-*entailed*: Each literal in $T$ is also in $\mu$.

**Lemma 3** *Let $\varphi$ be a true QBF. Let $\Pi$ be the minimal true subtree of the tree searched by Q-DLL$(\varphi, \epsilon)$. Let $\mu$ be an assignment for $\varphi$. If $\varphi_\mu$ is true, then the term associated with the node $\mu$ of $\Pi$*

- *is $\mu$-entailed; and*

- *does not contain universal literals assigned as monotone.*

**Proof.** Analogous to the proof of Lemma 2. □

**Theorem 9** *Let $\varphi$ be a true QBF. Let $\Pi$ the minimal true subtree of the tree searched by Q-DLL$(\varphi, \epsilon)$. Then $\Pi$ corresponds to a model generation and term resolution deduction of the empty term.*

**Proof.** Let $\Delta$ be the sequence of terms obtained by listing the terms associated with the nodes of $\Pi$ visited in post order. Clearly, $\Delta$ is a model generation and term resolution deduction. $\Delta$ is a deduction of the empty term because the node $\epsilon$ has an associated $\epsilon$-entailed term (Lemma 3), i.e., the empty term. □

As before, regardless of whether the input formula is true or false, the tree explored by *Q-DLL* may contain (exponentially many) subtrees whose nodes are associated with

---

1. For the sake of efficiency, it is also important that the term $T$ satisfies other properties. However, they are not necessary for the time being, and will be discussed in the next section.

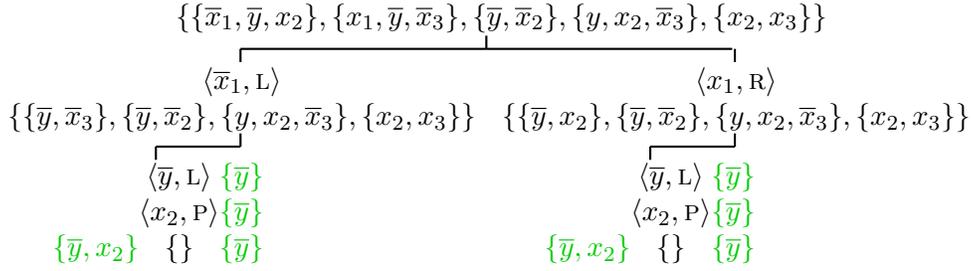

Figure 4: The term resolutions corresponding to the tree generated by *Q-DLL* for (4). The prefix is $\exists x_1 \forall y \exists x_2 \exists x_3$.

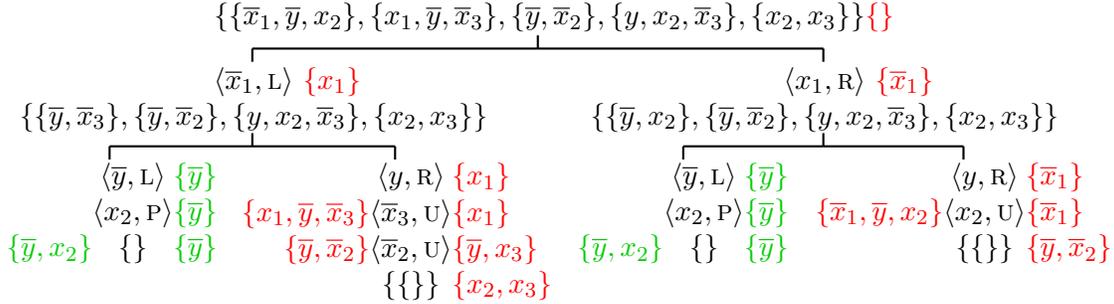

Figure 5: The resolution corresponding to the tree generated by *Q-DLL* for (4). The prefix is $\exists x_1 \forall y \exists x_2 \exists x_3$.

assignments $\mu$ with $\varphi_\mu$ being true. The above described procedure allows us to associate a term resolution deduction with each of such subtrees. For instance, if $\varphi$ is (4) there are two maximal such subtrees, having roots $\overline{x}_1; \overline{y}$ and $x_1; \overline{y}$. The associated deductions are represented in Figure 4. In the figure,

- we represent also the nodes along the path from the root to the subtrees,

- the term associated with each node is written in green and to the right of the node itself,

- if $\mu$ is a leaf, a non-contradictory term $T$ entailing the matrix and whose minimal form $min(T)$ is associated with $\mu$, is written in green and to the left of $\mu$.

Merging the trees in Figures 3 and 4 we obtain the whole tree of deductions corresponding to the search tree explored by *Q-DLL* (represented in Figure 5) in which clause and term resolutions are intermixed.

Now we consider the case in which the input QBF $\varphi$ is false and we no longer assume that literals are assigned as unit only if they are at the highest level. We restrict our attention to the minimal false subtree $\Pi$ of the tree searched by $Q\text{-}DLL(\varphi, \epsilon)$. Then, the procedure described above for associating a clause with each node of $\Pi$ may no longer work. For one thing, given a leaf $\mu$, there may be no $\mu$-falsified clauses in the matrix of the input formula. However, we are guaranteed about the existence of a $\mu$-contradicted clause in the matrix of the input formula. A clause $C$ is $\mu$-contradicted if[2]

- for each literal $l$ in $C$, $l$ is not in $\mu$; and

- for each existential literal $l$ in $C$, $\bar{l}$ is in $\mu$.

As long as we can associate with each node $\mu$ of $\Pi$ a $\mu$-contradicted clause (either belonging to the matrix of $\varphi$ or obtained by clause resolution) $\Pi$ corresponds to a clause resolution deduction of the empty clause: Indeed the clause associated with the root of $\Pi$ has to be empty (remember that the resolvent of a clause resolution is in minimal form). Thus, the obvious solution is to try to associate

1. with each leaf $\mu$ a $\mu$-contradicted clause in the input formula, and

2. with each internal node $\mu$ a $\mu$-contradicted clause obtained by resolving input clauses and/or previously deduced clauses along the same lines outlined before.

In some cases the process runs smoothly. Consider for instance, a QBF of the form:

$$\exists x_1 \exists x_2 \forall y \exists x_3 \{\{x_1, \overline{x}_3\}, \{\overline{x}_2, x_3\}, \{x_2, y, x_3\}, \ldots\}. \tag{7}$$

Then, if we assume that a split on $\overline{x}_1$ occurs first, the following path will be explored (we are using the same conventions of Figure 3):

$$\begin{array}{c} \langle \overline{x}_1, \text{L} \rangle \\ \langle \overline{x}_3, \text{U} \rangle \\ \langle \overline{x}_2, \text{U} \rangle \\ \{\{\}\} \end{array} \tag{8}$$

and the clause associated with each node are:

$$\begin{array}{ccc} & \langle \overline{x}_1, \text{L} \rangle & \color{red}{\{x_1\}} \\ \color{red}{\{x_1, \overline{x}_3\}} & \langle \overline{x}_3, \text{U} \rangle & \color{red}{\{x_1\}} \\ \color{red}{\{\overline{x}_2, x_3\}} & \langle \overline{x}_2, \text{U} \rangle & \color{red}{\{y, x_3\}} \\ & \{\{\}\} & \color{red}{\{x_2, y, x_3\}} \end{array}$$

where we see that:

1. the clause associated with the leaf $\mu = \overline{x}_1; \overline{x}_3; \overline{x}_2$ is not $\mu$-falsified but is $\mu$-contradicted; and

---

2. With respect to the definition of "contradictory clause" given in Section 3.2, it is clear that a clause $C$ is contradictory if and only if it is $\epsilon$-contradicted. Further, for any QBF $\varphi$ and assignment $\mu$, there exists a $\mu$-contradicted clause in the matrix of $\varphi$, if and only if $\varphi_\mu$ contains a $\epsilon$-contradicted clause, if and only if $\varphi_\mu$ contains a contradictory clause.

0 **function** *Rec-C-Resolve*$(\varphi, C_1, C_2, l, \mu)$
1    $S := \{l : l \in C_1, \bar{l} \in C_2\}$;
2    **if** $(S = \emptyset)$ **return** *C-Resolve*$(C_1, C_2)$;
3    $l_1 := \langle$an existential literal in $C_1$ with level $\leq$ than the level of all the literals in $C_1\rangle$;
4    $C := \langle$a clause in $\varphi$ which causes $\overline{l_1}$ to be unit in $\varphi_{\mu'}$, where $\mu'; \overline{l_1}$ is a prefix of $\mu\rangle$;
5    $C_3 := $ *C-Resolve*$(C_1, C)$;
6    **return** *Rec-C-Resolve*$(\varphi, C_3, C_2, l, \mu)$.

Figure 6: The algorithm of *Rec-C-Resolve*.

2. we are able to associate with each node $\mu$ a $\mu$-contradicted clause.

Unfortunately, in some cases things do not run so smoothly, i.e., it may not be possible to associate a clause to an internal node by a simple single resolution between input and/or previously deduced clauses. Indeed, some clause resolutions may be blocked because of universal variables occurring both as $y$ and $\bar{y}$ in the clauses to be used for the resolution. Consider for instance a QBF of the form (obtained from (7) by replacing the clause $\{\overline{x}_2, x_3\}$ with $\{\overline{x}_2, \overline{y}, x_3\}$):

$$\exists x_1 \exists x_2 \forall y \exists x_3 \{\{x_1, \overline{x}_3\}, \{\overline{x}_2, \overline{y}, x_3\}, \{x_2, y, x_3\}, \ldots\}. \tag{9}$$

Then, (8) would be still a valid path, and the corresponding clause resolutions would be:

$$\begin{array}{cc}
& \langle \overline{x}_1, \text{L} \rangle \\
\{x_1, \overline{x}_3\} & \langle \overline{x}_3, \text{U} \rangle \\
\{\overline{x}_2, \overline{y}, x_3\} & \langle \overline{x}_2, \text{U} \rangle \quad \ldots \\
& \{\{\}\} \quad \{x_2, y, x_3\}
\end{array} \tag{10}$$

where it is not possible to perform the clause resolution associated with the node having label $\langle \overline{x}_2, \text{U} \rangle$. As in the example, a clause resolution (5) may be blocked only because of some "blocking" universal literal $l$

- with both $l$ and $\bar{l}$ not in $\mu$, and

- with $l \in C_1$ and $\bar{l} \in C_2$.

Since both $C_1$ and $C_2$ are in minimal form, this is only possible if both $C_1$ and $C_2$ contain an existential literal $l'$

- having level less than or equal to the level of all the other literals in the clause; and

- assigned as unit.

Then, the obvious solution is to get rid, e.g., of the blocking literals $l$ in $C_1$ by resolving away from $C_1$ the existential literals with a level lower than the level of $l$.

This is the idea behind the procedure *Rec-C-Resolve* in Figure 6. In the figure, we assume that

1. $\varphi$ is the input QBF;

2. $\mu; l$ is an assignment;

3. $l$ is an existential literal which is either unit or at the highest level in $\varphi_\mu$;

4. $C_1$ is a clause containing $\bar{l}$, in minimal form and $\mu; l$-contradicted;

5. $C_2$ is a clause containing $l$, in minimal form and $\mu; \bar{l}$-contradicted. Further, if $l$ is unit in $\varphi_\mu$, then $C_2$ is a clause which causes $l$ to be unit in $\varphi_\mu$;

6. $C\text{-}Resolve(C_1, C_2)$ returns the resolvent of the clause resolution between the two clauses $C_1$ and $C_2$.

From here on, if $\langle \varphi, C_1, C_2, l, \mu \rangle$ satisfies the first 5 of the above conditions, we say that the pair $\langle C_1, C_2 \rangle$ *is to be* $\mu; l$-*Rec-C-Resolved (in* $\varphi$*)*. Given two clauses $\langle C_1, C_2 \rangle$ to be $\mu; l$-*Rec-C-Resolved*:

1. The set $S$ of universal literals blocking the clause resolution between $C_1$ and $C_2$ is computed (line 1).

2. If $S$ is empty, then we can simply return the resolvent between $C_1$ and $C_2$ (line 2); otherwise

3. we pick an existential literal $l_1$ in $C_1$ having minimum level in $C_1$ (line 3): $l_1$ has been assigned as unit earlier in the search, and we consider a clause $C$ which caused $l_1$ to be assigned as unit (line 4). If $C_3$ is the resolvent between $C_1$ and $C$ (line 5), $Rec\text{-}C\text{-}Resolve(\varphi, C_3, C_2, l, \mu)$ is returned (line 6).

If $\langle C_1, C_2 \rangle$ are to be $\mu; l$-*Rec-C-Resolved* in $\varphi$, $Rec\text{-}C\text{-}Resolve(\varphi, C_1, C_2, l, \mu)$ returns a minimal clause which is $\mu$-contradicted and without existential literals whose negation has been assigned as monotone in $\mu$. This is formally stated by the following lemma.

**Lemma 4** *Let $C_1$ and $C_2$ be two clauses such that $\langle C_1, C_2 \rangle$ is to be $\mu; l$-Rec-C-Resolved in a QBF $\varphi$. $Rec\text{-}C\text{-}Resolve(\varphi, C_1, C_2, l, \mu)$ terminates and returns a clause*

- *in minimal form and $\mu$-contradicted; and*

- *which does not contain existential literals whose negation has been assigned as monotone in $\mu$.*

The proof of the lemma is quite long and it is reported in the appendix.

Assuming that the input QBF $\varphi$ is false, the construction of the deduction of the empty clause (associated with the minimal false subtree $\Pi$ of the tree searched by *Q-DLL*) is the following:

- With every leaf $\mu$ of $\Pi$, we associate a clause $C$ in the input formula which is $\mu$-contradicted.

- If $C$ is the clause associated with a node $\mu; l$, then

  1. If $\bar{l} \notin C$ or if $l$ is universal then $C$ is the clause associated with the parent of $\mu; l$, i.e., with the node $\mu$. Notice that if $l$ is existential and monotone in $\varphi_\mu$ then $\bar{l} \notin C$.

2. If $\bar{l} \in C$ and $l$ is unit in $\varphi_\mu$ then the clause associated with the node $\mu$ is the result of *Rec-C-Resolve*$(\varphi, C, C', l, \mu)$, where $C'$ is a clause of $\varphi$ which causes $l$ to be unit in $\varphi_\mu$.

3. If $\bar{l} \in C$, $l$ is existential and not unit in $\varphi_\mu$, then we have to consider also the clause $C'$ associated with the node $\mu;\bar{l}$. If $l \notin C'$ then $C'$ is the clause associated with $\mu$ (as in the first case). If $l \in C'$, the clause associated with the node $\mu$ is the result of *Rec-C-Resolve*$(\varphi, C, C', l, \mu)$.

In our example, if $\varphi$ is (9) and with reference to the deduction in (10), the blocked resolution is the one associated with the node $\overline{x}_1; \overline{x}_3; \overline{x}_2$. *Rec-C-Resolve*$(\varphi, \{x_2, y, x_3\}, \{\overline{x}_2, \overline{y}, x_3\}, \overline{x}_2, \overline{x}_1; \overline{x}_3)$

1. at line 5, resolves $\{x_2, y, x_3\}$ and $\{x_1; \overline{x}_3\}$, and the resolvent $C_3$ is $min(\{x_1, x_2, y\}) = \{x_1, x_2\}$; and

2. the following recursive call to *Rec-C-Resolve*$(\varphi, \{x_1, x_2\}, \{\overline{x}_2, \overline{y}, x_3\}, \overline{x}_2, \overline{x}_1; \overline{x}_3)$ at line 6 returns $\{x_1, \overline{y}, x_3\}$.

Thus, the clause associated with each node are:

$$
\begin{array}{rcl}
 & \langle \overline{x}_1, \text{L} \rangle & \{x_1\} \\
\{x_1, \overline{x}_3\} & \langle \overline{x}_3, \text{U} \rangle & \{x_1\} \\
\{\overline{x}_2, \overline{y}, x_3\} & \langle \overline{x}_2, \text{U} \rangle & \{x_1, \overline{y}, x_3\} \\
 & \{\{\}\} & \{x_2, y, x_3\}
\end{array}
$$

Notice that, with reference to Figure 6, the choice of eliminating the blocking literals in $C_1$ while maintaining $C_2$ invariant, is arbitrary. Indeed, we could eliminate the blocking literals in $C_2$ and maintain $C_1$ invariant. In the case of the deduction in (10), this amounts to eliminate the universal literal $\overline{y}$ in $\{\overline{x}_2, \overline{y}, x_3\}$: By resolving this clause with $\{x_1, \overline{x}_3\}$ on $x_3$, we get the resolvent $\{x_1, \overline{x}_2\}$, which leads to the following legal deduction:

$$
\begin{array}{rcl}
 & \langle \overline{x}_1, \text{L} \rangle & \{x_1\} \\
 & \{x_1, \overline{x}_3\} & \langle \overline{x}_3, \text{U} \rangle & \{x_1\} \\
(\text{From } \{\overline{x}_2, \overline{y}, x_3\}, \{x_1, \overline{x}_3\}) & \{x_1, \overline{x}_2\} & \langle \overline{x}_2, \text{U} \rangle & \{x_1, y, x_3\} \\
 & \{\{\}\} & \{x_2, y, x_3\}
\end{array}
$$

**Lemma 5** *Let $\varphi$ be a false QBF. Let $\Pi$ be the minimal false subtree of the tree searched by Q-DLL$(\varphi, \epsilon)$. Let $\mu$ be an assignment for $\varphi$. If $\varphi_\mu$ is false, then the clause associated with the node $\mu$ of $\Pi$*

- *is in minimal form and $\mu$-contradicted; and*

- *does not contain existential literals whose negation has been assigned as monotone in $\mu$.*

**Proof.** By construction, each clause associated with a leaf $\mu$ of $\Pi$ is $\mu$-contradicted. We now show that also the clause $C$ associated with an internal node $\mu$ of $\Pi$ is $\mu$-contradicted, assuming that the clause $C'$ associated with its child $\mu; l$ is $\mu; l$-contradicted. If $\mu$ has also a child $\mu; \bar{l}$, we also assume that the clause $C''$ associated with its child $\mu; \bar{l}$ is $\mu; \bar{l}$-contradicted.

1. If $\bar{l} \notin C'$ or if $l$ is universal then $C = C'$. Hence, $C$ is in minimal form. Since $\bar{l} \notin C'$ or $l$ is universal, $C'$ is $\mu$; $l$-contradicted if and only if $C'$ is $\mu$-contradicted. The thesis follows because $C = C'$.

2. If $\bar{l} \in C'$ and $l$ is unit in $\varphi_\mu$ then $C =$ *Rec-C-Resolve*$(\varphi, C', C'', l, \mu)$, where $C''$ is a clause of $\varphi$ which causes $l$ to be unit in $\varphi_\mu$. The thesis follows from Lemma 4.

3. If $\bar{l} \in C$, $l$ is existential and not unit in $\varphi_\mu$, then we have to consider also the clause $C'$ associated with the node $\mu$; $\bar{l}$. Assuming $l \in C'$ (otherwise we would be in the first case), the clause associated with $\mu$ is the result of *Rec-C-Resolve*$(\varphi, C, C', l, \mu)$. As in the previous case, the thesis follows from Lemma 4.

□

**Theorem 10** *Let $\varphi$ be a false QBF. Let $\Pi$ be the minimal false subtree of the tree searched by Q-DLL$(\varphi, \epsilon)$. Then $\Pi$ corresponds to a clause resolution deduction of the empty clause.*

**Proof.** Given Lemma 5, the proof is analogous to the one of Theorem 7. □

## 4. Backjumping and Learning in *DLL* Based Procedures for QBFs

In this section we first show that computing the resolvent associated with each node allows to backjump some branches while backtracking (Subsection 4.1). Then, we show that learning resolvents allows to prune the search tree in branches different from the ones in which resolvents were computed and learned (Subsection 4.2).

### 4.1 Conflict and Solution Directed Backjumping

The procedure described in Section 3.2 uses a standard backtracking schema whenever the empty clause (resp. matrix) is generated: *Q-DLL* will backtrack up to the first existential (resp. universal) literal assigned as left split. For instance, given the QBF

$$\forall y_1 \exists x_1 \forall y_2 \exists x_2 \exists x_3 \{ \{y_1, y_2, x_2\}, \{y_1, \overline{y}_2, x_2, \overline{x}_3\}, \{y_1, \overline{x}_2, x_3\}, \\ \{\overline{y}_1, x_1, x_3\}, \{\overline{y}_1, y_2, x_2\}, \{\overline{y}_1, y_2, \overline{x}_2\}, \{\overline{y}_1, \overline{x}_1, \overline{y}_2, \overline{x}_3\} \}, \tag{11}$$

the tree searched by *Q-DLL* is represented in Figure 7, where we use the same conventions as in Section 3.

In the 2001 work of Giunchiglia, Narizzano, and Tacchella (2001), it is shown that the exploration of some branches is not necessary. In particular, if $\varphi$ is the input QBF and $\mu$ is an assignment, we show how it is possible to compute a "reason" for the (un)satisfiability of $\varphi_\mu$ while backtracking. Intuitively speaking, a reason for the result of $\varphi_\mu$ is a subset $\nu$ of the literals in $\mu$ such that for any other assignment $\mu'$

- which assigns to true or false the same literals assigned by $\mu$ (i.e., such that $\{|l| : l \text{ is in } \mu'\} = \{|l| : l \text{ is in } \mu\}$); and

- which extends $\nu$ (i.e., such that $\nu \subseteq \{l : l \text{ is in } \mu'\}$),

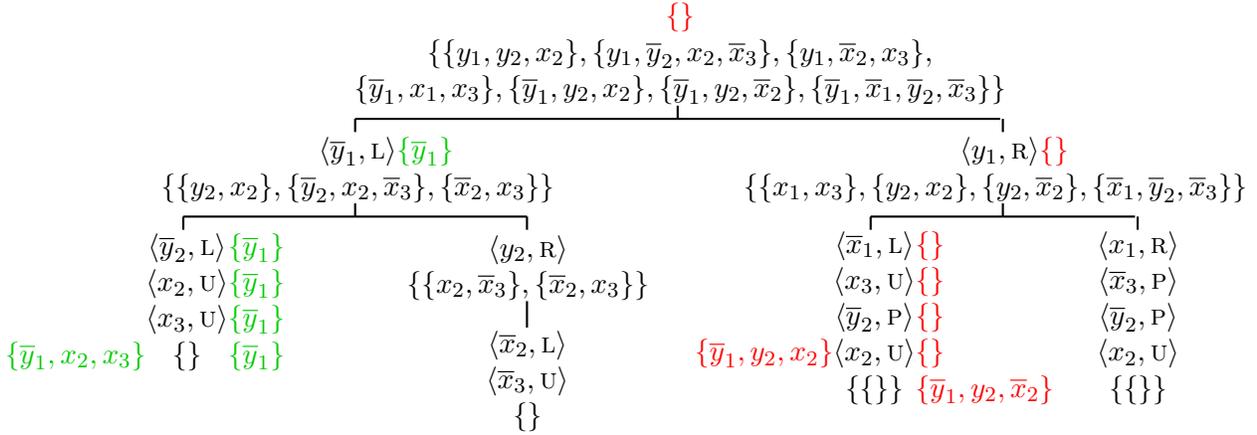

Figure 7: The resolution corresponding to the tree generated by *Q-DLL* for (11). The prefix is $\forall y_1 \exists x_1 \forall y_2 \exists x_2 \exists x_3$.

$\varphi_{\mu'}$ is equivalent to $\varphi_\mu$. Then, by computing reasons, we can avoid to right split on a literal $l$ if $\bar{l}$ is not in the reason: assigning $l$ to false would not change the result. The resulting procedure is a generalization to QBF of the popular Conflict-directed Backjumping (CBJ) (Prosser, 1993b), but also introduces the concept of Solution-directed Backjumping (SBJ), for avoiding useless splits on universal variables.

In a later paper, Giunchiglia, Narizzano, and Tacchella (2003) show how it is possible to optimize the computation of reasons. In particular, in that paper, it is shown that

- assuming $\varphi_\mu$ is unsatisfiable, we can consider reasons being a subset of the *existential* literals in $\mu$, while

- assuming $\varphi_\mu$ is satisfiable, we can consider reasons being a subset of the *universal* literals in $\mu$.

Apart from these optimizations, the tree searched by the procedures described in the former and latter papers is the same, and, in the case of (11), the exploration of the branches starting with $\langle y_2, \text{R} \rangle$, $\langle x_1, \text{R} \rangle$ will be skipped (see Figure 7).

We now show that the computation of the resolutions corresponding to *Q-DLL* allows to avoid the exploration of some branches pretty as much as CBJ and SBJ do: In the case of the QBF (11), the branches being skipped will be the same skipped by CBJ and SBJ.

The key point is to think about *Q-DLL* as a procedure producing a clause (resp. term) deduction of the empty clause (resp. term), proving that $\varphi$ is unsatisfiable (resp. satisfiable). Then, according to the rules we use for associating a deduction to the tree searched by *Q-DLL*, we have that:

- If $C$ is the clause associated with a node $\mu; l$ and $\bar{l} \notin C$, then the clause associated with the node $\mu$ is $C$, even if $l$ is existential and it has been assigned as left split.

```
0  function Q-DLL-BJ(φ, μ)
1     if (⟨a clause C is μ-contradicted⟩)
2        return C;
3     if (⟨the matrix of φμ is empty⟩) return ModelGenerate(μ);
4     if (⟨l unit in φμ⟩)
5        C := ⟨a clause in the matrix of φ which causes l to be unit in φμ⟩;
6        WR := Q-DLL-BJ(φ, μ; l);
7        if (⟨WR is a term⟩ or l̄ ∉ WR) return WR;
8        return Rec-C-Resolve(φ, WR, C, l, μ);
9     if (⟨l is monotone in φμ⟩) return Q-DLL-BJ(φ, μ; l);
10    l := ⟨a literal at the highest level in φμ⟩;
11    WR := Q-DLL-BJ(φ, μ; l);
12    if (⟨l is existential⟩ and (⟨WR is a term⟩ or l̄ ∉ WR)) return WR;
13    if (⟨l is universal⟩ and (⟨WR is a clause⟩ or l ∉ WR)) return WR;
14    WR' := Q-DLL-BJ(φ, μ; l̄);
15    if (⟨l is existential⟩ and (⟨WR' is a term⟩ or l̄ ∉ WR')) return WR';
16    if (⟨l is universal⟩ and (⟨WR' is a clause⟩ or l ∉ WR')) return WR';
17    if (⟨l is existential⟩) return Rec-C-Resolve(φ, WR', WR, l, μ);
18    return T-Resolve(WR', WR, l, μ).
```

Figure 8: The algorithm of *Q-DLL-BJ*.

- Analogously, if $T$ is the term associated with a node $\mu; l$ and $\bar{l} \notin T$, then the term associated with the node $\mu$ is $T$, even if $l$ is universal and it has been assigned as left split.

The above rules do not take into account the clause/term associated with the node $\mu; \bar{l}$, and thus there is no need to explore the branch starting with $\mu; \bar{l}$.

Consider for example Figure 7, in which we use the standard conventions and, e.g., write the clause (resp. term) associated with each node $\mu$ in red (resp. green) to the right of the node. With reference to the figure, it is clear that considering the term $\{\bar{y}_1\}$ associated with the node $\bar{y}_1; \bar{y}_2$, there is no need to explore the branch starting with $\langle y_2, \textsc{r} \rangle$ in order to associate a $\bar{y}_1$-entailed term with the node $\bar{y}_1$. Similarly, considering the empty clause $\{\}$ associated with the node $y_1; \bar{x}_1$, there is again no need to explore the branch starting with $\langle x_1, \textsc{r} \rangle$ in order to associate a $y_1$-contradicted clause with the node $y_1$.

The procedure *Q-DLL-BJ*$(φ, μ)$ in Figure 8 incorporates these ideas. In the figure,

- *ModelGenerate*$(μ)$ returns the minimal form of a non-contradictory and $μ$-entailed term $T$ such that

  - for each clause $C \in \Phi$, $C \cap T \neq \emptyset$; and

  - for each universal literal $l$ in $μ$ assigned as monotone, $l \notin T$.

- *Rec-C-Resolve*$(φ, C_1, C_2, l, μ)$ is as in Figure 6.

- *T-Resolve*$(T_1, T_2)$ returns the resolvent of the term resolution between the two terms $T_1$ and $T_2$.

The behavior *Q-DLL-BJ* can be illustrated in a few words by saying that *Q-DLL-BJ*$(\varphi, \mu)$ computes and returns the clause/term that would be associated with the node $\mu$ in the tree explored by *Q-DLL*. In particular, assuming

- that $WR$ is the clause (resp. term) returned by *Q-DLL-BJ*$(\varphi, \mu; l)$;

- that $l$ is existential (resp. universal); and

- that $l$ has been assigned as left split,

*Q-DLL-BJ*$(\varphi, \mu)$ does not explore the branch starting with $\mu; \overline{l}$ if $\overline{l} \notin WR$ (resp. $l \notin WR$), see line 12 (resp. line 13) in *Q-DLL-BJ*.

So far, with reference to Figure 7, we can interpret the clause (resp. term) in red (resp. green) to the right of a node $\mu$ as the value returned by *Q-DLL-BJ*$(\varphi, \mu)$. Then, considering the term $\{\overline{y}_1\}$ associated with the node $\overline{y}_1; \overline{y}_2$, *Q-DLL-BJ* does not explore the branch starting with $\langle y_2, \text{R} \rangle$. Similarly, considering the empty clause $\{\}$ associated with the node $y_1; \overline{x}_1$, *Q-DLL-BJ* again does not explore the branch starting with $\langle x_1, \text{R} \rangle$.

**Theorem 11** *Q-DLL-BJ$(\varphi, \epsilon)$ returns the empty clause if $\varphi$ is false, and the empty term if $\varphi$ is true.*

**Proof.**(Sketch) It is enough to notice that:

- If a node $\mu$ has associated a clause $C$, then $C$ is $\mu$-contradicted, and $C$ is the result of a sequence of clause resolutions.

- If a node $\mu$ has associated a term $T$, then $T$ is $\mu$-entailed, and $T$ is the result of a sequence of model generations and term resolutions.

Then, as in the previous section:

- If the empty clause is associated with the initial node $\epsilon$, then $\varphi$ is false.

- If the empty term is associated with the initial node $\epsilon$, then $\varphi$ is true. $\qquad\square$

### 4.2 Learning

Learning is a well known technique in SAT for avoiding the useless traversal of branches. In SAT, learning amounts to storing (clause) resolvents associated with the nodes of the tree explored by *DLL*: these resolvents are called "nogoods" and can be simply added to the set of input clauses.

In the case of QBFs, the situation is different and more complicated. Indeed, we have two types of resolutions ("term" and "clause"), and while the resolvents of clause resolutions can be added conjunctively to the matrix, the resolvents of term resolutions (that we will call "goods") have to be considered as in disjunction to the matrix.

In practice, we have to handle three sets of formulas:

- a set $\Psi$ of terms corresponding to the goods learned during the search;

- a set $\Phi$ of clauses corresponding to the matrix of the input QBF; and

- a set $\Theta$ of clauses corresponding to the nogoods learned during the search.

Formally, if $\varphi$ is a QBF of the form (1), a QBF $\varphi$ *Extended with Learning (EQBF)* is an expression of the form

$$Q_1 z_1 \ldots Q_n z_n \langle \Psi, \Phi, \Theta \rangle \qquad (n \geq 0) \tag{12}$$

where

- $\Psi$ is a set of terms, also called *goods*, to be interpreted disjunctively. Each good is obtained by model generation and/or term resolution from $\varphi$;

- $\Theta$ is a set of clauses, also called *nogoods*, to be interpreted conjunctively. Each nogood is obtained by clause resolution from $\varphi$.

Clearly,

$$Q_1 z_1 \ldots Q_n z_n (\Psi \vee \Phi)$$

and

$$Q_1 z_1 \ldots Q_n z_n (\Phi \wedge \Theta)$$

are equivalent to (1).

Initially $\Psi$ and $\Theta$ are the empty set, and $\Phi$ is the input set of clauses. As the search proceeds,

- Nogoods are determined while backtracking from a contradiction (i.e., on an assignment $\mu$ and $\varphi_\mu$ is unsatisfiable) and are possibly added to $\Theta$; and

- Goods are determined while backtracking from a solution (i.e., on an assignment $\mu$ and $\varphi_\mu$ is satisfiable) and are possibly added to $\Psi$.

In the following, we will use the term *constraints* when we want to refer to goods and nogoods indifferently.

Consider an EQBF (12). Because of the constraints in $\Psi$ and/or $\Theta$, the search can be pruned considerably. Indeed, while descending the search tree, any literal can be assigned as long as we are guaranteed that we can reconstruct a valid clause/term deduction –while backtracking– of the empty clause/term. The availability of already derived clauses/terms allows to prune the search because of the constraints in $\Psi$ or $\Theta$: Given an assignment $\mu$, if there exists a $\mu$-contradicted clause $C \in \Theta$ (resp. a $\mu$-satisfied term $T \in \Psi$) we can stop the search and return $C$ (resp. $T$). A term $T$ is $\mu$-*satisfied* if

- for each literal $l$ in $T$, $\bar{l}$ is not in $\mu$; and

- for each universal literal $l$ in $T$, $l$ is in $\mu$.

Clearly, a $\mu$-entailed term is also $\mu$-satisfied. Further, we can extend the notion of unit to take into account the constraints in $\Psi$ and/or $\Theta$. A literal $l$ is

- *unit* in a EQBF (12) if

```
0  function Rec-Resolve(ψ, W₁, W₂, l, μ)
1      S := {l : l ∈ W₁, l̄ ∈ W₂};
2      if (S = ∅) return Resolve(W₁, W₂);
3      l' := ⟨a literal in W₁ with level ≤ than the level of all the literals in W₁⟩;
4      W := ⟨a constraint in ψ which causes l' to be unit in ψ_μ', where μ'; l' is a prefix of μ⟩;
5      W₃ := Resolve(W₁, W);
6      return Rec-Resolve(ψ, W₃, W₂, l, μ).
```

Figure 9: The algorithm of *Rec-Resolve*.

- either $l$ is existential and for some $m \geq 0$,

  * a clause $\{l, l_1, \ldots, l_m\}$ belongs to $\Phi$ or $\Theta$, and
  * each expression $\forall |l_i|$ $(1 \leq i \leq m)$ occurs at the right of $\exists |l|$ in the prefix of (12).

- or $l$ is universal and for some $m \geq 0$,

  * a term $\{l̄, l_1, \ldots, l_m\}$ belongs to $\Psi$, and
  * each expression $\exists |l_i|$ $(1 \leq i \leq m)$ occurs at the right of $\forall |l|$ in the prefix of (12).

As for the definition of monotone literals, the crucial property that has to be ensured when dealing with EQBFs, is that an existential (resp. universal) literal $l$ assigned as monotone in $\mu; l$ should never enter in a nogood (resp. good) associated with a node extending $\mu; l$. This is guaranteed by defining a literal $l$ as *monotone* or *pure* if and only if[3]

- either $l$ is existential and $l̄$ does not belong to any constraint in $\Phi \cup \Theta$;

- or $l$ is universal and $l$ does not belong to any constraint in $\Psi \cup \Phi$.

Because of the possibility of assigning also universal literals as unit, it may be the case that some term resolutions may be blocked because of some existential literals $l$ and $l̄$, each occurring in one of the terms to be used in the antecedents of the term resolution. However, the procedure *Rec-C-Resolve* presented in in Subsection 3.3 can be easily generalized to work also for the case in which the constraints to be resolved are terms. The result is the procedure *Rec-Resolve*($\psi, W_1, W_2, l, \mu$) in Figure 9, where it is assumed that

1. $\psi$ is an EQBF;

---

3. There are various ways to guarantee that an existential literal $l$ assigned as monotone in $\mu; l$ does not enter in a nogood associated with a node extending $\mu; l$. Another one is to

- keep the definition of existential monotone literal unchanged: An existential literal can be assigned as monotone in (12) if $l̄$ does not belong to any clause in $\Phi$; and

- update $\Theta$ to (or proceed in the search as if $\Theta$ has been updated to) $\Theta \setminus \{C : C \in \Theta, l̄ \in C\}$.

Analogously for universal monotone literals. See the work of Giunchiglia, Narizzano and Tacchella (2004a) for more details and possibilities, including a discussion about the interaction between the monotone rule and learning.

2. $\mu; l$ is an assignment;

3. $l$ is an existential (resp. universal) literal which is either unit or at the highest level in $\psi_\mu$;

4. $W_1$ is a clause (resp. term) containing $\bar{l}$ (resp. $l$), in minimal form and $\mu; l$-contradicted (resp. $\mu; l$-satisfied);

5. $W_2$ is a clause (resp. term) containing $l$ (resp. $\bar{l}$), in minimal form and $\mu; \bar{l}$-contradicted (resp. $\mu; \bar{l}$-satisfied). Further, if $l$ is unit in $\psi_\mu$, then $W_2$ is a clause (resp. term) which causes $l$ to be unit in $\psi_\mu$;

6. for each existential (resp. universal) literal $l'$ assigned as unit in $\mu'; l'$, with $\mu'; l'$ a prefix of $\mu; l$, there has to be a clause (resp. term) in $\psi$ which causes $l'$ to be unit in $\psi_{\mu'}$.

7. $Resolve(W_1, W_2)$ returns $C\text{-}Resolve(W_1, W_2)$ (resp. $T\text{-}Resolve(W_1, W_2)$).

If $\langle \psi, W_1, W_2, l, \mu \rangle$ satisfy the first 6 of the above 7 conditions, we say that the pair $\langle W_1, W_2 \rangle$ *is to be $\mu; l$-Rec-Resolved (in $\psi$)*.

In the above, if $\psi$ is (12), $\psi_l$ is defined as the EQBF obtained from $\psi$ by

- removing from $\Phi$ and $\Theta$ (resp. $\Psi$) the clauses $C$ (resp. terms $T$) with $l \in C$ (resp. $\bar{l} \in T$), and by removing $\bar{l}$ (resp. $l$) from the other clauses in $\Phi \cup \Theta$ (resp. terms in $\Psi$); and

- removing $Q|l|$ from the prefix.

If $\mu = l_1; l_2; \ldots; l_m$ $(m \geq 0)$, $\psi_\mu$ is defined as $(\ldots ((\psi_{l_1})_{l_2}) \ldots)_{l_m}$.

If $\langle W_1, W_2 \rangle$ are to be $\mu; l$-Rec-Resolved in $\psi$, $Rec\text{-}Resolve(\psi, W_1, W_2, l, \mu)$ returns a constraint in minimal form and $\mu$-contradicted or $\mu$-satisfied, as stated by the following lemma.

**Lemma 6** *Let $W_1$ and $W_2$ be two clauses (resp. terms) such that $\langle W_1, W_2 \rangle$ is to be $\mu; l$-Rec-Resolved in a EQBF $\psi$. $Rec\text{-}Resolve(\psi, W_1, W_2, l, \mu)$ terminates and returns a minimal clause (resp. term) which*

- *is $\mu$-contradicted (resp. $\mu$-satisfied); and*

- *does not contain existential literals whose negation has been (resp. universal literals which have been) assigned as monotone in $\mu$.*

**Proof.** (Sketch) The proof is equal to (resp. analogous to) the proof of Lemma 4 if $l$ is existential (resp. universal). □

The procedure $Q\text{-}DLL\text{-}LN(\varphi, \mu)$ incorporates the above new definitions and ideas, and is represented in Figure 10. Considering the figure,

- the definition of $ModelGenerate(\mu)$ can be relaxed with respect to the definition provided in Subsection 4.1 in order to return the minimal form of a non-contradictory and $\mu$-satisfied term $T$ such that

```
0   Ψ := {};
1   Θ := {};
2   function Q-DLL-LN(φ, μ)
3       Q := ⟨the prefix of φ⟩;
4       Φ := ⟨the matrix of φ⟩;
5       if (⟨a μ-contradicted clause C is in Φ ∪ Θ⟩)
6           return C;
7       if (⟨a μ-satisfied term T is in Ψ⟩)
8           return T;
9       if (⟨the matrix of φ_μ is empty⟩) return ModelGenerate(μ);
10      if (⟨l is unit in (Q⟨Ψ, Φ, Θ⟩)_μ⟩)
11          W := ⟨a constraint in Ψ ∪ Φ ∪ Θ which causes l to be unit in (Q⟨Ψ, Φ, Θ⟩)_μ⟩;
12          WR := Q-DLL-LN(φ, μ; l);
13          if (⟨l is existential⟩ and (⟨WR is a term⟩ or l̄ ∉ WR)) return WR;
14          if (⟨l is universal⟩ and (⟨WR is a clause⟩ or l ∉ WR)) return WR;
15          WR := Rec-Resolve(Q⟨Ψ, Φ, Θ⟩, WR, W, l, μ);
16          Learn(μ, WR);
17          return WR;
18      if (⟨l is monotone in (Q⟨Ψ, Φ, Θ⟩)_μ⟩) return Q-DLL-LN(φ, μ; l);
19      l := ⟨a literal at the highest level in φ_μ⟩;
20      WR := Q-DLL-LN(φ, μ; l);
21      if (⟨l is existential⟩ and (⟨WR is a term⟩ or l̄ ∉ WR)) return WR;
22      if (⟨l is universal⟩ and (⟨WR is a clause⟩ or l ∉ WR)) return WR;
23      WR' := Q-DLL-LN(φ, μ; l̄);
24      if (⟨l is existential⟩ and (⟨WR' is a term⟩ or l̄ ∉ WR')) return WR';
25      if (⟨l is universal⟩ and (⟨WR' is a clause⟩ or l ∉ WR')) return WR';
26      WR := Rec-Resolve(Q⟨Ψ, Φ, Θ⟩, WR', WR, l, μ);
27      Learn(μ, WR);
28      return WR.
```

Figure 10: The algorithm of $Q$-DLL-LN.

  – for each clause $C \in \Phi$, $C \cap T \neq \emptyset$; and

  – for each universal literal $l$ in $\mu$ assigned as monotone, $l \notin T$.

• $Learn(\mu, WR)$ updates the set of goods and nogoods according to a given policy. Here we simply assume that $Learn(\mu, WR)$ updates $\Psi$ and $\Theta$ to $\Psi'$ and $\Theta'$ respectively, and that $\Psi'$ and $\Theta'$ satisfy the following conditions:

  – $\Psi'$ is a subset of $\Psi \cup \{WR\}$ if $WR$ is a term, and of $\Psi$ otherwise;

  – $\Theta'$ is a subset of $\Theta \cup \{WR\}$ if $WR$ is a clause, and of $\Theta$ otherwise; and

  – for each existential (resp. universal) literal $l$ assigned as unit in an initial prefix $\mu'; l$ of $\mu$, $\Theta' \cup \Phi$ (resp. $\Psi'$) still contains a clause (resp. term) that causes $l$ to be assigned as unit in $(Q\langle\Psi', \Phi, \Theta'\rangle)_{\mu'}$.

With reference to Figure 9, this last condition is necessary in order to guarantee the existence of a constraint $W$ satisfying the condition at line 4.

The above conditions on $Learn(\mu, WR)$ are very general and ensure the soundness and completeness of $Q$-DLL-LN.

**Theorem 12** $Q$-DLL-LN$(\varphi, \epsilon)$ returns the empty clause if $\varphi$ is false, and the empty term if $\varphi$ is true.

**Proof.** Analogous to the proof of Theorem 11. □

To understand the benefits of learning, assume the input QBF is (4). The corresponding EQBF is

$$\exists x_1 \forall y \exists x_2 \exists x_3 \langle \{\}, \{\{\overline{x}_1, \overline{y}, x_2\}, \{\overline{y}, \overline{x}_2\}, \{x_2, x_3\}, \{x_1, \overline{y}, \overline{x}_3\}, \{y, x_2, \overline{x}_3\}\}, \{\}\rangle,$$

and the search proceeds as in Figure 2, with the first path leading to the empty matrix, which starts the term resolution process. Assuming the term $min(\{\overline{y}, x_2\}) = \{\overline{y}\}$ is added to the set of goods before checking the value of $\varphi_{\overline{x}_1;y}$, as soon as $x_1$ is assigned to true,

- $y$ is detected to be unit and it is correspondingly assigned; and

- the path corresponding to the assignment $x_1; \overline{y}$ is not explored.

As this example shows, (good) learning can avoid the useless exploration of some branches that would be explored with a backtracking or backjumping schema. Indeed, we have been assuming that the deduced term is learned while backtracking. A policy according to which $Learn(\mu, WR)$ simply adds $WR$

- to $\Theta$ if $WR$ is clause; and

- to $\Psi$ otherwise,

can be easily implemented. However, such simple policy may easily lead to store an exponential number of goods and/or nogoods (notice that we have a call to $Learn(\mu, WR)$ for each literal assigned as unit or right split). Thus, practical implementations incorporate policies guaranteed to be space bounded, i.e., ones that store a polynomial number of goods and nogoods at most. In SAT, the three most popular space bounded learning schemes are:

- *Size learning of order $n$* (Dechter, 1990): a nogood is added to $\Theta$ if and only if its cardinality is less or equal to $n$. Once added, it is never deleted.

- *Relevance learning of order $n$* (Ginsberg, 1993): given a current assignment $\mu$, a nogood $C$ is always added to $\Theta$, and then it is deleted from $\Theta$ as soon as the number of literals $l$ in $C$ and with $\bar{l} \notin \mu$ is bigger than $n$.

- *Unique Implication Point (UIP) based learning* (Marques-Silva & Sakallah, 1996): a nogood $C$ is stored if and only if $C$ contains only one literal at the maximum *decision level*. Given an assignment $\mu$, the *decision level* of a literal $l$ in $\mu$ is the number of splits done before $l$ in $\mu$. With UIP based learning, the set $\Theta$ of added clauses is periodically inspected and clauses are deleted according to various criteria.

Thus, in size learning, once a nogood is stored, it is never deleted. In relevance and UIP based learning, nogoods are dynamically added and deleted depending on the current assignment. See the work of Bayardo (1996) for more details related to size and relevance learning (including their complexity analysis), and the work of Zhang, Madigan, Moskewicz and Malik (2001) for a discussion of various UIP based learning mechanisms for SAT. Size, relevance, UIP based learning are just a few of the various possibilities for limiting the number of stored clauses, and each one can be generalized in various ways when considering QBFs instead of SAT formulas. In the next section, we will present the particular learning schema that we implemented in QuBE.

## 5. Implementation and Experimental Analysis

In this section we first describe in some details the implementation of nogood and good learning in QuBE, and then we report on some experimental analysis conducted in order to evaluate the (separate) benefits of nogood and good learning, but also the relative efficiency of our solver when compared to other state-of-the-art QBF solvers.

### 5.1 Implementation in QuBE

To evaluate the benefits deriving from learning, we have implemented both good and nogood learning in QuBE. QuBE is a QBF solver based on search which, on non-random instances, compares well with respect to other state-of-the-art solvers based on search, like SEMPROP (Letz, 2002), YQUAFFLE (Zhang & Malik, 2002a), i.e., the best solvers based on search on non-random instances according to (Le Berre, Simon, & Tacchella, 2003), see (Giunchiglia, Narizzano, & Tacchella, 2004c) for more details.

Besides learning, the version of QuBE that we used features

- efficient detection of unit and monotone literals using lazy data structures as in (Gent, Giunchiglia, Narizzano, Rowley, & Tacchella, 2004);

- a branching strategy that exploits information gleaned from the input formula initially, and leverages the information extracted in the learning phase.

See (Giunchiglia, Narizzano, & Tacchella, 2004b) for a description of these characteristics.

As for learning, the computation of nogoods and goods corresponding to the internal nodes of the search tree is carried out by doing clause and term resolution between a "working reason" which is initialized when backtracking starts, and the "reasons" stored while descending the search tree

- for each unit literal, the stored reason is a constraint in which the literal is unit;

- for each literal assigned as right split, the stored reason is the constraint computed while backtracking on the left branch;

- for monotone literals, the way working reasons are initialized ensures that existential (resp. universal) monotone literals never belong to a working reason computed while backtracking from a contradiction (resp. solution).

Assume that $\mu = l_1; l_2; \ldots; l_m$ is the assignment corresponding to the leaf under consideration. Considering the problem of initializing the working reason, the way we do it in QuBE is to

- return a $\mu$-contradicted clause in the matrix of the input QBF or in the set of learned nogoods, if we have a contradiction; and

- compute the minimal form of a $\mu$-satisfied prime implicant of the matrix which contains as few universal literals as possible, if we have a solution.

In the second case, the computation of a prime implicant is important in order to have short reasons, while having as few as possible universal literals is important in order to backjump nodes. The above requirements are met by recursively removing irrelevant literals from the set of literals in $\mu$, starting from the universals ones. Given a set $S$ of literals, we say that a literal is *irrelevant in $S$* if for each clause $C$ in the matrix with $l \in C$ there exists another literal $l'$ in $S$ with $l' \in C$. If $prime(\mu)$ is the set of literals being the result of the recursive procedure, the term $\bigwedge prime(\mu)$

- is satisfied by $\mu$;

- is a prime implicant of the matrix of the input QBF;

- is such that there does not exist another term satisfying the first two properties and with a smaller (under set inclusion) set of universal literals.

In order to further reduce the number of universal literals in the initial goods, we take advantage of the fact that the assignment $\mu$ may be partial: For some literal $l$ it may be the case that neither $l$ nor $\bar{l}$ is in $\mu$. Then, we can use the existential literals not in $\mu$, and with level lower than the level of all the universal literals in $\mu$ assigned as left split, in order to further reduce the number of universals in $prime(\mu)$. In fact, for any sequence $\mu'$ of literals extending $\mu$ with existential literals, the set of universals in $prime(\mu')$ is a subset of $prime(\mu)$. For instance, considering the QBF (11) if $\mu = \bar{y}_1; y_2; \bar{x}_2; \bar{x}_3$, then

- $prime(\mu)$ is $\{\bar{y}_1, y_2, \bar{x}_2, \bar{x}_3\}$; and

- if we extend $\mu$ to $\mu' = \mu; x_1$ then $prime(\mu')$ is $\{x_1, y_2, \bar{x}_2, \bar{x}_3\}$.

Finally, when evaluating which universal literals in $\mu$ are irrelevant, we follow the reverse order in which they have been assigned, in order to try to backjump as high as possible in the search tree.

As we said in the previous section, besides the problem of setting the initial working reason, another problem with learning is that unconstrained storage of clauses (resp. terms) obtained by the reasons of conflicts (resp. solutions) may lead to an exponential memory blow up. In practice, it is necessary to introduce criteria

1. for limiting the constraints that have to be learned; and/or

2. for unlearning some of them.

The implementation of learning in QuBE works as follows. Assume that we are backtracking on a literal $l$ assigned at decision level $n$. The constraint corresponding to the reason for the current conflict (resp. solution) is learned only if the following conditions are satisfied:

1. $l$ is existential (resp. universal); and

2. all the assigned literals in the reason except $l$, are at a decision level strictly lower than $n$; and

3. there are no open universal (resp. existential) literals in the reason that are before $l$ in the prefix.

Notice that these three conditions ensure that $l$ is unit in the constraint corresponding to the reason. Once QuBE has learned the constraint, it backjumps to the node at the maximum decision level among the literals in the reason, excluding $l$. We say that $l$ is a *Unique Implication Point* (UIP) and therefore the lookback in QuBE is "UIP based". Notice that our definition of UIP generalizes to QBF the concepts first described by Silva and Sakallah (1996) and used in the SAT solver GRASP. On a SAT instance, QuBE lookback scheme behaves similarly to the "1-UIP-learning" scheme used in zCHAFF (and described in Zhang et al., 2001). Even if QuBE is guaranteed to learn at most one clause (resp. term) per each conflict (resp. solution), still the number of learned constraints may blow up, as the number of backtracks can be exponential. To stop this course, QuBE scans periodically the set of learned constraints in search of those that became irrelevant, i.e., clauses (resp. terms) where the number of open literals exceeds a parameter $n$, corresponding to the relevance order. Thus, our implementation uses UIP based learning to decide when to store a constraint, and a relevance based criteria to decide when to forget a constraint. In the experimental analysis presented in the next subsection, the parameter $n$ has been set to 20 and the set of learned constraints is scanned every 5000 nodes.

Besides the above learning mechanism, our current version of QuBE features lazy data structures for unit literal detection and propagation (as described in Gent et al., 2004), monotone literal fixing (as described in Giunchiglia et al., 2004a), and a Variable State Independent Decaying Sum heuristic (VSIDS) (as introduced in SAT by Moskewicz, Madigan, Zhao, Zhang, & Malik, 2001). As in SAT, the basic ideas of our heuristic are to (*i*) initially rank literals on the basis of the occurrences in the matrix, (*ii*) increment the weight of the literals in the learned constraints, and (*iii*) periodically divide by a constant the weight of each literal.

## 5.2 Experimental Results

To evaluate the effectiveness of our implementation, we considered the 450 formal verification and planning benchmarks that constituted part of the 2003 QBF solvers comparative evaluation[4]: 25% of these instances comes from verification problems (described in Scholl & Becker, 2001; Abdelwaheb & Basin, 2000), and the remaining are from planning domains (described in Rintanen, 1999; Castellini, Giunchiglia, & Tacchella, 2001). We start our analysis considering QuBE with and without learning enabled. Both versions of QuBE compute goods and nogoods in order to backjump over irrelevant existential and universal branching nodes. They differ in the treatment of the computed goods and nogoods:

- when learning is enabled, QuBE records both goods and nogoods;

- when learning is disabled, QuBE records neither nogoods nor goods.

---

4. With respect to the non-random instances used in the 2003 QBF comparative evaluation, our test set does not include the QBF encodings of the modal K formulas submitted by Pan and Vardi (2003).

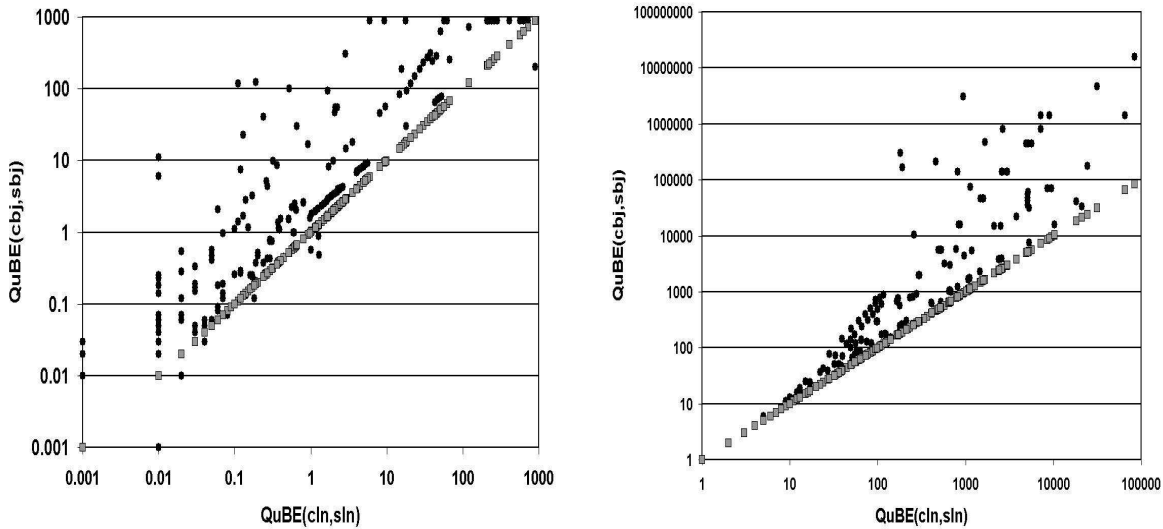

Figure 11: Effectiveness of learning: QuBE versus QuBE(cbj,sbj). CPU time (left) and number of backtracks on the instances solved by both solvers (right).

We call the two versions QuBE(cln,sln) and QuBE(cbj,sbj) respectively, in order to specify the type of look-back used by the two systems. Notice that we did not consider QuBE with backtracking (i.e., the version which computes neither nogoods nor goods and performs simple chronological backtracking) because it is not competitive with the other solvers.

All the experiments were run on a farm of identical PCs, each one equipped with a Pentium 4, 3.2GHz processor, 1GB of RAM, running `Linux Debian (sarge)`. Finally, each system had a timeout value of 900s per instance.

Figure 11 left shows the performances of QuBE(cln,sln) versus QuBE(cbj,sbj). In the plot, the $x$-axis is the CPU-time of QuBE(cln,sln) and the $y$-axis is the CPU-time of QuBE(cbj,sbj). A plotted point $\langle x, y \rangle$ represents a benchmark on which QuBE(cln,sln) and QuBE(cbj,sbj) take $x$ and $y$ seconds respectively.[5] For convenience, we also plot the points $\langle x, x \rangle$, each representing the benchmarks solved by QuBE(cln,sln) in $x$ seconds. The first observation is that learning pays off:

---

5. In principle, one point $\langle x, y \rangle$ could correspond to many benchmarks solved by QuBE(cln,sln) and QuBE(cbj,sbj) in $x$ and $y$ seconds respectively. However, in this and in the other scatter diagrams that we present, each point (except for the point $\langle 900, 900 \rangle$, representing the instances on which both solvers time-out) corresponds to a single instance in most cases.

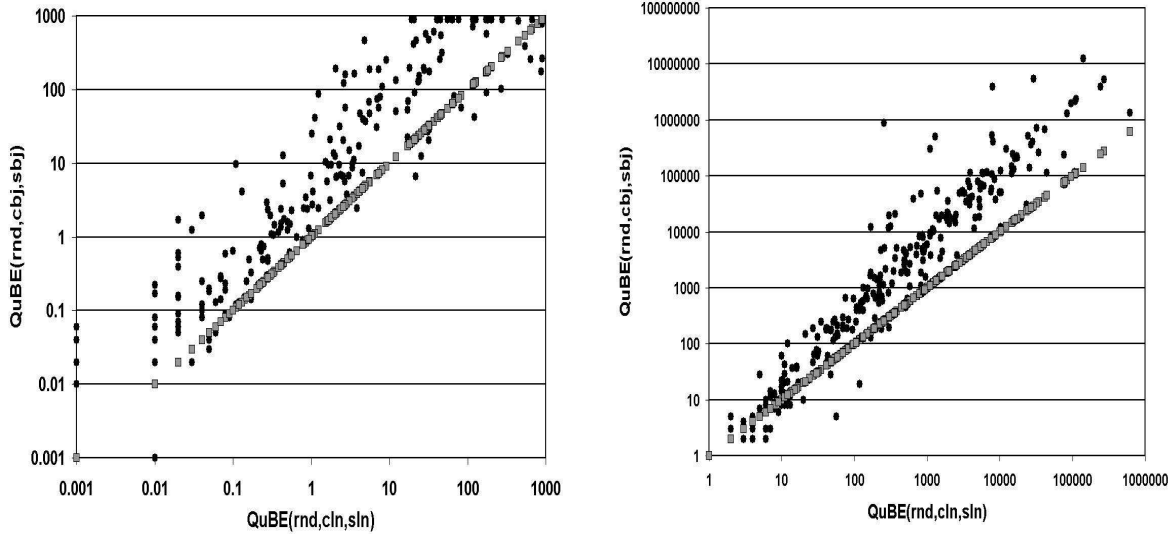

Figure 12: Effectiveness of learning with a random heuristic: QuBE(RND,CLN,SLN)[3] versus QuBE(RND,CBJ,SBJ)[3]. CPU time (left) and number of backtracks on the instances solved by both solvers (right).

- QuBE(CLN,SLN) (resp. QuBE(CBJ,SBJ)) is able to solve 16 (resp. 1) instances that are not solved by QuBE(CBJ,SBJ) (resp. QuBE(CLN,SLN)); and

- among the instances solved by both solvers, QuBE(CLN,SLN) (resp. QuBE(CBJ,SBJ)) is at least one order of magnitude faster than QuBE(CBJ,SBJ) (resp. QuBE(CLN,SLN)) on 39 (resp. 0) instances.

In order to have an implementation-quality independent measure of the pruning introduced by learning, the right plot in the figure shows the number of backtracks (i.e., the number of solutions and conflicts found) of QuBE(CBJ,SBJ) versus QuBE(CLN,SLN) on the 358 problems solved by both systems. Here a plotted point $\langle x, y \rangle$ represents a benchmark that is solved by QuBE(CLN,SLN) and QuBE(CBJ,SBJ) performing $x$ and $y$ backtracks respectively. As it can be seen, learning substantially prunes the search space: There is no point below the diagonal, meaning that it is never the case that QuBE(CBJ,SBJ) performs less backtracks than QuBE(CLN,SLN).[6] Still, learning has some overhead, and thus the pruning caused by learning not always pays off in terms of speed, as proved by the few points below the diagonal in the left plot.

The above experimental data are not entirely satisfactory for two reasons.

First, learning and the heuristic are tightly coupled in QuBE: Whenever QuBE learns a constraint, it also increments the score of the literals in it. In QuBE(CBJ,SBJ) no constraint is ever learned. As a consequence, in QuBE(CBJ,SBJ), (*i*) literals are initially sorted on the basis of their occurrences in the input QBF, and (*ii*) the score of each literal is periodically halved until it becomes 0. When all the literals have score 0, then literals at the same prefix level are chosen according to their lexicographic order.

---

6. This does not imply that the tree searched by QuBE(CLN,SLN) is a subtree of the tree searched by QuBE(CBJ,SBJ): Indeed, the literal selected at each branching node by the two systems is not guaranteed to be the same.

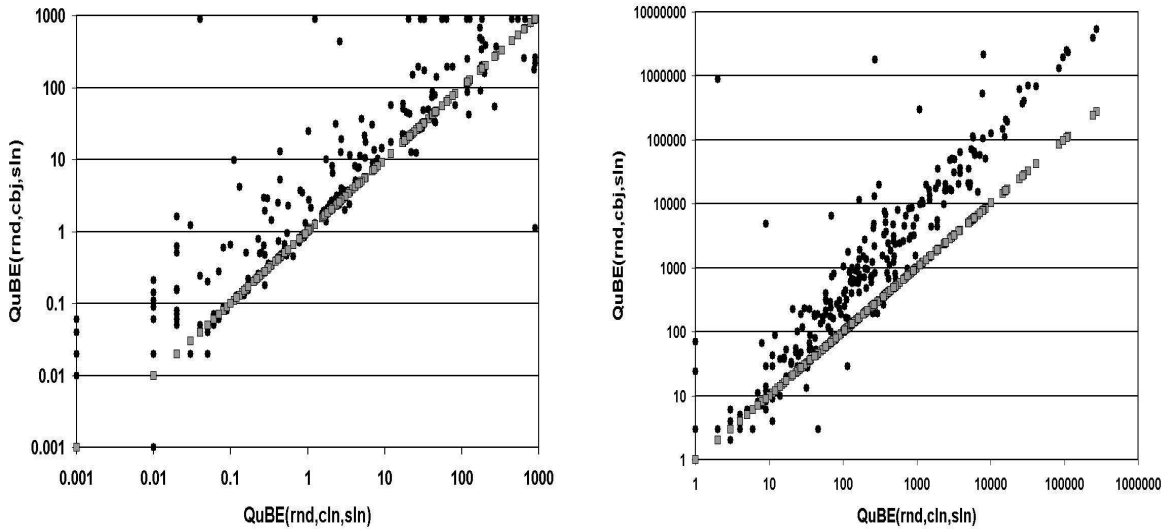

Figure 13: Effectiveness of conflict learning: QuBE(RND,CLN,SLN)[3] versus QuBE(RND,CBJ,SLN)[3]. CPU time (left) and number of conflict backtracks on the instances solved by both solvers (right).

Second, independently from the heuristic being used, a plot showing the performances of QuBE with and without learning, does not say which of the two learning schemes (conflict, solution) is effective (Gent & Rowley, 2004).

To address the first problem, we consider QuBE with a random heuristic, i.e., a heuristic which randomly selects a literal among those at the maximum level and not yet assigned. We call the resulting systems QuBE(RND,CLN,SLN) and QuBE(RND,CBJ,SBJ) respectively: As the names suggest, the first has learning enabled, while in the second learning has been disabled. Because of the randomness, we run each solver 5 times on each instance. Then, we define QuBE(RND,CLN,SLN)[*i*] to be the system whose performances are, on a each instance, the *i*-th best among the 5 results obtained by running QuBE(RND,CLN,SLN) on that instance. QuBE(RND,CBJ,SBJ)[*i*] is defined analogously.

Figure 12 shows the CPU time (left) and number of backtracks on the solved instances (right) of QuBE(RND,CLN,SLN)[3] and QuBE(RND,CBJ,SBJ)[3]. From the plots, it is easy to see that QuBE(RND,CLN,SLN)[3] is faster than QuBE(RND,CBJ,SBJ)[3] in most cases. To witness this fact

- QuBE(RND,CLN,SLN) (resp. QuBE(RND,CBJ,SBJ)) is able to solve 21 (resp. 2) instances that are not solved by QuBE(CBJ,SBJ) (resp. QuBE); and

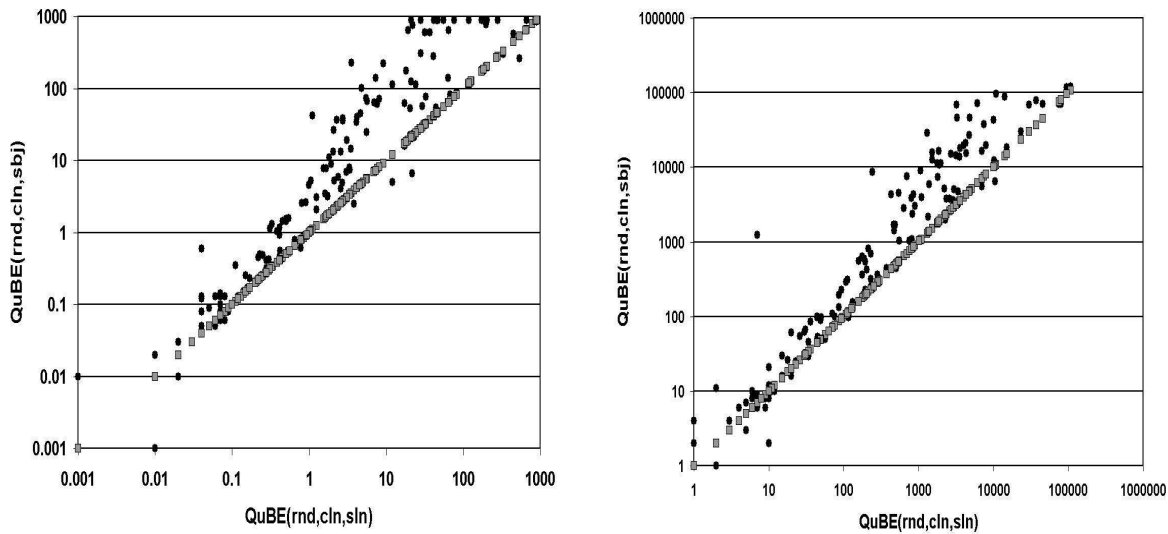

Figure 14: Effectiveness of solution learning: QuBE(RND,CLN,SLN)[3] versus QuBE(RND,CLN,SBJ)[3]. CPU time (left) and number of solution backtracks on the instances solved by both solvers (right).

- among the instances solved by both solvers, QuBE (resp. QuBE(CBJ,SBJ)) is at least one order of magnitude faster than QuBE(CBJ,SBJ) (resp. QuBE) on 68 (resp. 2) instances.

Still, it is no longer the case that enabling learning always causes a reduction in the number of backtracks. This can be because of the different literals selected at each branching node, but also because pruning a node may prevent a "long backjump" (Prosser, 1993a) which would cause a vast reduction of the search space. Interestingly, comparing with the results in Figure 11, it seems that with a random heuristic learning becomes more important. This fact witnesses also in our setting the well known tension between look-ahead and look-back techniques: A "smart" look-ahead makes the look-back less important, and viceversa.

To address the second problem, we considered the systems QuBE(RND,CBJ,SLN) and QuBE(RND,CLN,SBJ), i.e., the systems obtained from QuBE(RND,CLN,SLN) by disabling conflict learning and solution learning respectively. As usual, each system was run 5 times on each instance, and QuBE(RND,CBJ,SLN)[i] and QuBE(RND,CLN,SBJ)[i] ($1 \leq i \leq 5$) are defined as before. The left plots in Figures 13 and 14 show the performances of QuBE(RND,CLN,SLN)[3] versus QuBE(RND,CBJ,SLN)[3] and QuBE(RND,CLN,SBJ)[3] respectively. We also measured the number of backtracks. However, in order to better highlight the pruning due to conflict (resp. solution) learning, the right plot in Figure 13 (resp. 14) shows the number of conflict (resp. solution) backtracks of QuBE(RND,CBJ,SLN)[3] (resp. QuBE(RND,CLN,SBJ)[3]). From the plots, we see that both conflict and solution learning prune the search space and pay off: In each plot, there are only a few points well below the diagonal. Comparing the two left plots, we also see that, on the test set that we considered, solution learning helps in solving problems more than conflict learning: QuBE(RND,CBJ,SLN)[3] times out on 101 while QuBE(RND,CLN,SBJ)[3] times out on 107. On the other hand, the two right plots suggest that conflict learning prunes more than solution learning, but this conclusion is not correct. Indeed, each plot shows either the number of conflicts or the number of solutions: Pruning a node (no matter whether existential or universal) may avoid finding (exponentially many) solutions and/or conflicts. In particular, given that all the instances are in CNF and thus have the form

$$\ldots \forall y \exists x_1 \exists x_2 \ldots \exists x_n \Phi$$

| QuBE(RND,CLN,SLN)[3] | = | < | > | ≪ | ≫ | ⋈ | ×10< | ×0.1> | TO |
|---|---|---|---|---|---|---|---|---|---|
| QuBE(RND,CLN,SLN)[1] | 136 | 0 | 225 | 0 | 3 | 86 | 0 | 43 | 86 |
| QuBE(RND,CLN,SLN)[2] | 169 | 0 | 192 | 0 | 1 | 88 | 0 | 19 | 88 |
| QuBE(RND,CLN,SLN)[4] | 156 | 203 | 0 | 2 | 0 | 89 | 27 | 0 | 91 |
| QuBE(RND,CLN,SLN)[5] | 109 | 244 | 0 | 8 | 0 | 89 | 61 | 0 | 97 |
| QuBE(RND,CBJ,SBJ)[1] | 131 | 145 | 72 | 13 | 7 | 82 | 27 | 20 | 95 |
| QuBE(RND,CBJ,SBJ)[2] | 137 | 164 | 43 | 17 | 2 | 87 | 43 | 7 | 104 |
| QuBE(RND,CBJ,SBJ)[3] | 123 | 192 | 25 | 21 | 2 | 87 | 68 | 2 | 108 |
| QuBE(RND,CBJ,SBJ)[4] | 110 | 205 | 17 | 29 | 2 | 87 | 83 | 1 | 116 |
| QuBE(RND,CBJ,SBJ)[5] | 84 | 222 | 10 | 45 | 2 | 87 | 99 | 1 | 132 |
| QuBE(RND,CBJ,SLN)[1] | 130 | 96 | 128 | 7 | 5 | 84 | 20 | 26 | 91 |
| QuBE(RND,CBJ,SLN)[2] | 133 | 134 | 82 | 12 | 5 | 84 | 27 | 14 | 96 |
| QuBE(RND,CBJ,SLN)[3] | 129 | 169 | 48 | 15 | 3 | 86 | 40 | 5 | 101 |
| QuBE(RND,CBJ,SLN)[4] | 115 | 209 | 20 | 17 | 1 | 88 | 54 | 1 | 105 |
| QuBE(RND,CBJ,SLN)[5] | 86 | 245 | 6 | 24 | 1 | 88 | 87 | 0 | 112 |
| QuBE(RND,CLN,SBJ)[1] | 135 | 78 | 142 | 6 | 4 | 85 | 7 | 36 | 91 |
| QuBE(RND,CLN,SBJ)[2] | 151 | 110 | 90 | 10 | 4 | 85 | 15 | 15 | 95 |
| QuBE(RND,CLN,SBJ)[3] | 169 | 134 | 39 | 19 | 1 | 88 | 29 | 5 | 107 |
| QuBE(RND,CLN,SBJ)[4] | 141 | 183 | 11 | 26 | 0 | 89 | 51 | 0 | 115 |
| QuBE(RND,CLN,SBJ)[5] | 103 | 218 | 2 | 38 | 0 | 89 | 69 | 0 | 127 |

Table 2: Comparison among various versions of QuBE. Each row compares a system written in the first column with respect to QuBE(RND,CLN,SLN)[3] taken as reference. If $A$ is QuBE(RND,CLN,SLN)[3] and $B$ is a solver in the first column, then the other columns report the number of problems that: "=", $A$ and $B$ solve in the same time; "<", $A$ and $B$ solve but $A$ takes less time than $B$; ">", $A$ and $B$ solve but $A$ takes more time than $B$; "≪", $A$ solves while $B$ does not; "≫", $A$ does not solve while $B$ does; "⋈", $A$ and $B$ do not solve; "×10<", both $A$ and $B$ solve but on which $A$ is at least one order of magnitude faster; "×0.1<", both $A$ and $B$ solve but on which $A$ is at least one order of magnitude slower; "TO", $B$ does not solve. The number of timeouts for QuBE(RND,CLN,SLN)[3] is 89.

($n \geq 1$) pruning any variable not in $\{x_1, x_2, \ldots, x_n\}$ has the potential to prune $2^n$ conflicts.

Some further and more detailed quantitative information about the CPU times is reported in Table 2. From the last column in the table we see that, if we indicate with $TO(S)$ the number of timeouts of system $S$, then, for each $i \in \{1, 2, 3, 4, 5\}$,

$$TO(\text{QuBE}(\text{RND,CLN,SLN})[i]) < \frac{TO(\text{QuBE}(\text{RND,CBJ,SLN})[i])}{TO(\text{QuBE}(\text{RND,CLN,SBJ})[i])} < TO(\text{QuBE}(\text{RND,CBJ,SBJ})[i]).$$

The above gives an indication of the "capacity" of the solvers, i.e., of their ability to solve problems. In order to get an indication of their "productivity", i.e., considering the problems that they solve, their ability to solve them quickly, we can consider the number $FS(S)$ being the difference between the "$\times 0.1 >$" and "$10\times <$" columns: The lower $FS(S)$ is, the better $S$ is. Here again we have

$$FS(\text{QuBE}(\text{RND,CLN,SLN})[i]) < \frac{FS(\text{QuBE}(\text{RND,CBJ,SLN})[i])}{FS(\text{QuBE}(\text{RND,CLN,SBJ})[i])} < FS(\text{QuBE}(\text{RND,CBJ,SBJ})[i])$$

for $i \in \{1, 2, 3, 4, 5\}$. From the above, it is clear that both conflict and solution learning allow to improve on capacity and productivity. Our experimental results thus seem to contradict the negative results reported in Gent's and Rowley's work (2004) for solution based look-back mechanisms. However, those results are not comparable with ours, given the different mechanisms implemented by the respective solvers (e.g., for computing the initial solution and for monotone literal fixing), and the different experimental setting (e.g., the testset).

## 6. Conclusions and Related Work

This paper is based on and extends (Giunchiglia et al., 2002) which introduces nogood and good learning for QBFs satisfiability. Here we show the correspondence between the computation trees searched by DLL based QBF solvers and clause/term resolution deductions. Nogoods and goods are the clauses and terms respectively in the resolution deductions. Under this perspective, learning simply amounts to storing nogoods and goods. We show how to incorporate nogoods and goods learning in DLL based QBF solvers by considering EQBFs (QBFs extended with learning), and then illustrate by means of examples that the computation of nogoods and goods:

- allows for solution and conflict directed backjumping in the spirit of (Giunchiglia et al., 2001, 2003); and

- if stored, allows for pruning branches in other parts of the search tree.

We present a high level description of algorithms incorporating such ideas, and formally prove their soundness and completeness. We also discuss the problems related to effective implementations in DLL based QBF solvers, and present (in some details) our implementation in QuBE, a state-of-the-art QBF solver. The experimental analysis shows that QuBE enhanced with nogood and good learning is more effective, when considering a selection of nonrandom problems consisting of planning and formal verification benchmarks. We also show that QuBE is competitive with respect to the state of the art.

As we already said, our work builds on (Giunchiglia et al., 2002). Other papers dealing with learning in QBFs satisfiability are (Letz, 2002), (Zhang & Malik, 2002a) and (Gent & Rowley, 2004). In particular, in (Letz, 2002) conflict and solution learning are called lemma and model caching. The paper also proposes a technique based on model caching for dealing with QBFs having variable-independent subformulas. Zhang and Malik (2002a) propose conflict learning (which is then extended to solution learning in Zhang & Malik, 2002b). In the second paper, terms are called cubes. Gent and Rowley (2004) introduce a new form of solution learning: This new technique revisits less solutions than standard techniques, but the experimental results reported in the paper are not positive. All these works share the same intuitions and thus propose similar techniques. Though it is difficult to establish a precise relation among these works due to the differences in the terminology and/or the different level of detail in the presentations,[7] we believe that the main differences are at the implementation level, i.e., in the way solution and conflict learning have been implemented. It is therefore quite difficult if not impossible to compare the different alternatives, without re-implementing or recasting the different learning mechanisms or even the different solvers in a common framework. Indeed, the specific learning mechanism implemented within a solver may be motivated by the other characteristics of the solver, e.g., by the data structures being used or by the heuristic. For instance, watched data structures (used, e.g., in QuBE, yquaffle but not in semprop) allow for more efficient detection and propagation of unit and pure literals (Gent et al., 2004). As a consequence, solvers with watched data structures may profitably maintain huge databases for goods and nogoods. For solvers with standard data structures, the costs involved in managing such huge databases may overwhelm the advantages. Considering each of these solvers as a whole, the experimental analysis conducted in (Giunchiglia et al., 2004c) shows that our solver QuBE compares well with respect to semprop and yquaffle on the 450 formal verification and planning benchmarks that we considered also in this paper.

## Acknowledgments

We would like to thank Ian Gent and Andrew Rowley for discussions related to the subject of this paper, and the anonymous reviewers for their suggestions and corrections. This work has been partially supported by MIUR.

## Appendix A. Proof of Lemma 4

The proof is by well founded induction. Thus, the steps we follow are:

1. the definition of a well founded order on tuples $\langle C_1, C_2, l, \mu \rangle$;

2. the proof that the thesis holds for the minimal elements of the partial order; and

3. assuming that the thesis holds for all the tuples $\langle C_3, C_2, l, \mu \rangle$ such that $\langle C_3, C_2, l, \mu \rangle \preceq \langle C_1, C_2, l, \mu \rangle$, the proof that the thesis holds also for $\langle C_1, C_2, l, \mu \rangle$.

---

7. For instance, in (Letz, 2002; Zhang & Malik, 2002b) but also in our initial work (Giunchiglia et al., 2002), the method used for computing the initial working reason corresponding to a solution (procedure *ModelGenerate* in Figure 10) is not detailed.

We have deliberately omitted what are the properties that the elements of the tuples $\langle C_1, C_2, l, \mu \rangle$ in the partial order have to satisfy. Indeed, the standard assumption would be that $C_1$ and $C_2$ be two clauses such that $\langle C_1, C_2 \rangle$ is to be $\mu; l$-Rec-C-Resolved. However, this is not sufficient. Indeed, it may happen that starting from two clauses $\langle C_1, C_2 \rangle$ to be $\mu; l$-Rec-C-Resolved (line numbers refer to Figure 6)

1. the set $\{l : l \in C_1, \bar{l} \in C_2, l \text{ is universal}\}$ is not empty (see line 1);

2. the clause $C_3$ computed as in line 5 of Figure 6 is not $\mu; l$-contradicted; and thus

3. the tuple $\langle C_3, C_2, l, \mu \rangle$ is not an element of the partial order.

To better understand the problem, consider the following simple example:

$$\exists x_1 \forall y_1 \exists x_2 \forall y_2 \exists x_3 \exists x_4 \{\{\bar{x}_4\}, \{x_2, y_2, x_4\}, \{\bar{y}_2, x_3\}, \{\bar{x}_1, \bar{y}_1, \bar{x}_3\}, \{x_1, y_1, \bar{x}_2, \bar{x}_3\}\}. \tag{13}$$

For this QBF $\varphi$:

1. $\bar{x}_4; x_2; y_2; x_3; \bar{x}_1$ is an assignment producing a contradictory clause;

2. $\langle \{x_1, y_1, \bar{x}_2, \bar{x}_3\}, \{\bar{x}_1, \bar{y}_1, \bar{x}_3\} \rangle$ are to be $\bar{x}_4; x_2; y_2; x_3; \bar{x}_1$-Rec-C-Resolved;

3. Rec-C-Resolve$(\varphi, \{x_1, y_1, \bar{x}_2, \bar{x}_3\}, \{\bar{x}_1, \bar{y}_1, \bar{x}_3\}, \bar{x}_1, \bar{x}_4; x_2; y_2; x_3; \bar{x}_1)$,

    (a) causes a call to Rec-C-Resolve$(\varphi, \{x_1, y_1, y_2, x_4\}, \{\bar{x}_1, \bar{y}_1, \bar{x}_3\}, \bar{x}_1, \bar{x}_4; x_2; y_2; x_3; \bar{x}_1)$, in which the clause $C_3 = \{x_1, y_1, y_2, x_4\}$ is not $\bar{x}_4; x_2; y_2; x_3; \bar{x}_1$-contradicted; but

    (b) returns the clause $\{\bar{y}_1, \bar{x}_3\}$ which is $\bar{x}_4; x_2; y_2; x_3; \bar{x}_1$-contradicted, as expected.

The fact that the universal literal $y_2$ which causes $C_3$ not to be $\bar{x}_4; x_2; y_2; x_3; \bar{x}_1$-contradicted does not appear in the clause returned by Rec-C-Resolve is due to the following two facts:

1. $y_2$ has a lower level than the blocking literal $y_1$; and

2. the negation of all the existential literals in $C_3$ with a level lower than $y_2$ are assigned before $y_2$ in $\bar{x}_4; x_2; y_2; x_3; \bar{x}_1$.

To formally define these notions, we need some additional notation. First, consider a given clause $C_2$. $Res_{C_2}(C_1)$ is the set of literals in $C_1$ with a level lower than a literal blocking the resolution between $C_1$ and $C_2$. Formally:

$$Res_{C_2}(C_1) = \{l : l \in C_1, \exists l' \in Blocking_{C_2}(C_1) level(l) < level(l')\},$$

where

- for each literal $l$, $level(l)$ is the prefix level of $l$; and

- $Blocking_{C_2}()$ is the function defined by

$$Blocking_{C_2}(C_1) = \{l : l \in C_1, \bar{l} \in C_2, l \text{ is universal}\}$$

Let $\mu$ be an assignment. We say that a clause $C_1$ is $\mu$-contradictable (with respect to $C_2$) if

1. for each existential literal $l$ in $C_1$, $\bar{l}$ is in $\mu$;

2. for each universal literal $l$ in $C_1$, if $l$ is in $\mu$ then

   (a) $l \in Res_{C_2}(C_1)$; and

   (b) for each existential literal $l'$ in $C_1$, if $level(l') < level(l)$ then $\overline{l'}$ is to the left of $l$ in $\mu$.

Clearly, if a clause is $\mu$-contradicted then it is also $\mu$-contradictable. Considering the QBF (13), the clause $\{x_1, y_1, y_2, x_4\}$ is not $\overline{x}_4; x_2; y_2; x_3; \overline{x}_1$-contradicted; but is $\overline{x}_4; x_2; y_2; x_3; \overline{x}_1$-contradictable (with respect to $\{\overline{x}_1, \overline{y}_1, \overline{x}_3\}$).

Our well founded order and induction will be on the set of tuples $\langle C_1, C_2, l, \mu \rangle$ in which $C_1$ is $\mu; l$-contradictable. As a preliminary step, we first define the well founded order on literals according to which $l' \preceq l''$ if and only if either $l' = l''$ or both $\overline{l'}$ and $\overline{l''}$ are in $\mu$ and $\overline{l'}$ has been assigned before $\overline{l''}$ in $\mu$ (i.e., $\overline{l'}$ is to the left of $\overline{l''}$ in $\mu$).

We extend the partial order relation $\preceq$ from literals to clauses $(i)$ in minimal form, $(ii)$ containing $\bar{l}$, and $(iii)$ $\mu; l$-contradictable, by saying that for two such clauses $C_1$ and $C_3$, $C_3 \preceq C_1$ if

- either $C_3 = C_1$;

- or $\exists l' \in (Res^E_{C_2}(C_1) \backslash Res^E_{C_2}(C_3)) \forall l'' \in (Res^E_{C_2}(C_3) \backslash Res^E_{C_2}(C_1)) l' \preceq l''$, where $Res^E_{C_2}(C_1)$ is the subset of existential literals in $Res_{C_2}(C_1)$, and similarly for $Res^E_{C_2}(C_3)$.

The above order is well founded, and the minimal elements are such that $Res_{C_2}(C)$ (or, equivalently, $Blocking_{C_2}(C)$) is empty.

Finally, consider the set $W$ of tuples $\langle C_1, C_2, l, \mu \rangle$ such that

1. $\mu; l$ is an assignment;

2. $l$ is an existential literal which is either unit or at the highest level in $\varphi_\mu$;

3. $C_1$ is a clause containing $\bar{l}$, in minimal form and $\mu; l$-contradictable with respect to $C_2$;

4. $C_2$ contains $l$, is in minimal form and is $\mu; \bar{l}$-contradicted. Further, if $l$ is unit in $\varphi_\mu$, then $C_2$ is a clause which causes $l$ to be unit in $\varphi_\mu$.

On such set, we define a well founded order according to which $\langle C_3, C_2, l, \mu \rangle \preceq \langle C_1, C_2, l, \mu \rangle$ if $C_3 \preceq C_1$.

Now consider the procedure *Rec-C-Resolve* in Figure 6. We prove by well founded induction that, for each tuple $\langle C_1, C_2, l, \mu \rangle \in W$, *Rec-C-Resolve*$(\varphi, C_1, C_2, l, \mu)$ terminates and returns a clause $C$ in minimal form and $\mu$-contradicted. At the end, we will also show that if we further assume that $C_1$ is $\mu; l$-contradicted (and not simply $\mu; l$-contradictable), then $C$ does not contain existential literals whose negation has been assigned as monotone in $\mu$.

In the base case, $C_1$ is such that $Res_{C_2}(C_1)$ is empty. Hence, for each universal literal $l \in C_1$, $l$ is not in $\mu$ and thus $C_1$ is $\mu; l$-contradicted. Since $Res_{C_2}(C_1)$ is empty, the set $S$ computed at line 1 is empty and thus *Rec-C-Resolve*$(\varphi, C_1, C_2, l, \mu)$ terminates returning the resolvent $C$ of $C_1$ and $C_2$. Clearly $C$ is in minimal form, and it is easy to show that $C$ is $\mu$-contradicted.

For the step case, by induction hypothesis, we have that the thesis holds for *Rec-C-Resolve*$(\varphi, C_3, C_2, l, \mu)$ and we have to show that it holds also for *Rec-C-Resolve*$(\varphi, C_1, C_2, l, \mu)$, assuming $\langle C_3, C_2, l, \mu \rangle \preceq \langle C_1, C_2, l, \mu \rangle$. If the set $Res_{C_2}(C_1)$ is empty, then see the base case. Assume that $Res_{C_2}(C_1)$ is not empty, and thus that also $Blocking_{C_2}(C_1)$ is not empty.

From here on, let $l'$ be a literal in $Blocking_{C_2}(C_1)$ with the highest level. $l'$ is not in $\mu$ because $l' \notin Res_{C_2}(C_1)$ and $C_1$ is $\mu; l$-contradictable. $\overline{l'}$ is not in $\mu$ because $\overline{l'} \in C_2$ and $C_2$ is $\mu; l$-contradicted. Further, $level(l') < level(l)$. To see why, consider the only two possible cases:

1. $l$ is unit in $\varphi_\mu$: Since $\overline{l'} \in C_2$, and $C_2$ is a clause which causes $l$ to be unit in $\varphi_\mu$, it must be $level(l') = level(\overline{l'}) < level(l)$.

2. $l$ is at the highest level in $\varphi_\mu$: Since both $l'$ and $\overline{l'}$ are not in $\mu$ and $l$ is at the highest level in $\varphi_\mu$, $level(l') \leq level(l)$. On the other hand, $level(l') \neq level(l)$ because $l'$ is universal and $l$ is existential.

Since $C_1$ is in minimal form, there exists an existential literal $l''$ such that $l'' \in C_1$, $\overline{l''}$ is in $\mu$, and with $level(l'') < level(l') < level(l)$. From here on, let $l_1$ be an existential literal in $C_1$ (not necessarily distinct from $l''$) with level less than or equal to the level of all the literals in $C_1$ (see line 3). Since

$$level(l_1) < level(l') < level(l) \tag{14}$$

and $\overline{l_1}$ is in $\mu$ (because $C_1$ is $\mu; l$-contradictable), it follows that $\overline{l_1}$ has been assigned as unit, and thus there exists a clause $C$ in $\varphi$ which causes $\overline{l_1}$ to be unit in $\varphi_{\mu'}$, where $\mu'; \overline{l_1}$ is an initial prefix of $\mu$ (see line 4).

Consider the set

$$Blocking_C(C_1) = \{l : l \in C_1, \overline{l} \in C, l \text{ is universal}\}.$$

$Blocking_C(C_1)$ is empty. In fact, for each universal literal $l'' \in C$

- if $level(l'') < level(l_1)$ then $\overline{l''} \notin C_1$ since $C_1$ is in minimal form;

- if $level(l'') > level(l_1)$ then $\overline{l''} \preceq l_1$. Assume that $\overline{l''} \in C_1$. Since $C_1$ is $\mu; l$-contradictable, $l_1 \preceq \overline{l''}$. However, $\overline{l''} \preceq l_1$ and $l_1 \preceq \overline{l''}$ is not possible because $l_1 \neq \overline{l''}$ ($l_1$ is existential and $l''$ is universal).

Since $Blocking_C(C_1)$ is empty, we can resolve $C$ and $C_1$ on $l_1$, obtaining

$$C_3 = min((C_1 \cup C) \setminus \{l_1, \overline{l_1}\})$$

as resolvent. $C_3$ is in minimal form and it contains $\overline{l}$.

To show that $C_3 \preceq C_1$ it remains to be showed that $C_3$ is $\mu; l$-contradictable. Indeed, for each existential literal $l$ in $C_3$, $\overline{l}$ is in $\mu$, while for the universal literals in $C_3$, consider the two cases:

1. $Blocking_{C_2}(C_3)$ is not empty. In this case, $l' \in Blocking_{C_2}(C_3)$. This is an easy consequence of the following facts:

    (a) for each literal $l'' \in Blocking_{C_2}(C)$, $level(l'') < level(l')$: $\overline{l''} \in C_2$ by definition of $Blocking_{C_2}(C)$, hence $\overline{l''}$ is not in $\mu$ because $C_2$ is $\mu;l$-contradicted, and therefore $level(l'') < level(l_1)$, and thus the thesis (see (14));

    (b) $Blocking_{C_2}(C_3) = (Blocking_{C_2}(C_1) \cup Blocking_{C_2}(C)) \cap C_3$ and thus $(Blocking_{C_2}(C_1) \cup Blocking_{C_2}(C)) \setminus Blocking_{C_2}(C_3) = (C \cup C_1) \setminus (\{l_1, \overline{l_1}\} \cup C_3)$, i.e., the literals in $Blocking_{C_2}(C_1) \cup Blocking_{C_2}(C)$ and not in $Blocking_{C_2}(C_3)$ are those that have been omitted because of the minimal form of $C_3$;

    (c) $Blocking_{C_2}(C_3)$ is not empty.

    Since $l' \in Blocking_{C_2}(C_3)$, the literals in $Res_{C_2}(C_1)$ which are also in $C_3$, also belong to $Res_{C_2}(C_3)$, i.e.,

    $$Res_{C_2}(C_3) \supseteq Res_{C_2}(C_1) \cap C_3. \qquad (15)$$

    Now consider a universal literal $l'' \in C_1 \cap C_3$. If $l''$ is in $\mu$ then

    (a) $l'' \in Res_{C_2}(C_1)$ because $C_1$ is $\mu;l$-contradictable, and hence $l'' \in Res_{C_2}(C_3)$ (see (15));

    (b) for each existential literal $l'''$ in $C_1$, if $level(l''') < level(l'')$ then $\overline{l'''} \preceq l''$ because $C_1$ is $\mu;l$-contradictable;

    (c) for each existential literal $l''' \neq l_1$ in $C$, $\overline{l'''} \preceq l''$. In fact, $level(l_1) < level(l'')$, $l_1 \preceq l''$ because $C_1$ is $\mu;l$-contradictable, and for each existential literal $l''' \neq l_1 \in C$, $\overline{l'''} \preceq l_1$.

    Finally, consider a universal literal $l'' \in C \cap C_3$. If $l''$ is in $\mu$ then $level(l'') < level(l_1)$ and hence

    (a) $l'' \in Res_{C_2}(C_3)$ because $level(l_1) < level(l')$ (see (14)); and

    (b) for each existential literal $l''' \in C_3$, if $level(l''') < level(l'')$ then $l''' \in C$ and hence $\overline{l'''} \preceq l''$.

2. $Blocking_{C_2}(C_3)$ is empty. Let $m$ be the lowest among the level of the literals in $C_3$. $level(l') < m$ since $l' \notin C_3$. Then, for each universal literal $l'' \in C_3$, $l''$ is not in $\mu$, i.e., $C_3$ is $\mu;l$-contradicted. In fact, assume that there exists a universal literal $l'' \in C_3$ in $\mu$. Then, $level(l'') > m$ and either $l'' \in C_1$ or $l'' \in C$. Consider the first case $l'' \in C_1$. Then, $l'' \in Res_{C_2}(C_1)$ because $C_1$ is $\mu;l$-contradictable, and then $level(l'') < level(l')$. But this is not possible because $level(l'') > m$ and $level(l') < m$. Consider the case $l'' \in C$. Then, $level(l'') < level(l_1)$ and hence $level(l'') < level(l')$ (see (14)) which is again not possible.

Since $C_3 \preceq C_1$, $\langle C_3, C_2, l, \mu \rangle \preceq \langle C_1, C_2, l, \mu \rangle$, we can conclude by induction hypothesis that $Rec\text{-}C\text{-}Resolve(\varphi, C_3, C_2, l, \mu)$ returns a clause in minimal form and $\mu$-contradicted.

Now we make the further assumption that the input clause $C_1$ is $\mu; l$-contradicted. Then, $C_1$ does not contain existential literals whose negation has been assigned as monotone, and the same holds for $C_2$ and for each clause $C$ used at line 5. Hence, *Rec-C-Resolve*$(\varphi, C_3, C_2, l, \mu)$ returns a clause without existential literals whose negation has been assigned as monotone in $\mu$.